%% file: CoFi-Dec.tex
\documentclass[sigconf]{acmart}

\AtBeginDocument{%
  }

\setcopyright{acmlicensed}
\copyrightyear{2018}
\acmYear{2018}
\acmDOI{10.1145/3746027.3754791}

\usepackage{subfig}
\usepackage{amsmath}
\usepackage{mathtools}
\usepackage{amsthm}
\usepackage{multirow}
\usepackage{colortbl}
\usepackage{adjustbox}

\usepackage{soul}
\usepackage{lipsum}
\usepackage{fancyhdr}

\usepackage{eccvabbrv}
\usepackage{graphicx}
\usepackage[ruled]{algorithm2e}
\usepackage{upgreek}
\usepackage{wrapfig}
\usepackage{xcolor}
\usepackage{pifont}
\usepackage{graphicx}
\usepackage{multirow}
\newtheorem{theorem}{Theorem}[section]

\theoremstyle{definition}

\newtheorem{remark}[theorem]{Remark}
\usepackage[ruled]{algorithm2e}
\definecolor{cvprblue}{rgb}{0.21,0.49,0.74}
\definecolor{greenx}{RGB}{0,128,128}
\definecolor{maroonx}{RGB}{195,18,48}
\usepackage{tcolorbox}
\usepackage{wrapfig}
\usepackage{textgreek}
\usepackage{bbm}

\newcommand{\cc}{\cellcolor{gray!20}}
\definecolor{darkred}{RGB}{192, 0, 0}
\definecolor{darkgreen}{RGB}{103, 174, 64}

\newcommand{\RebuttalRevision}[1]{\textcolor{black}{#1}}

\setcopyright{acmlicensed}
\acmDOI{10.1145/3746027.3754791}
\acmISBN{979-8-4007-2035-2/2025/10}

\copyrightyear{2025}
\acmYear{2025}
\setcopyright{acmlicensed}\acmConference[MM'25]{Proceedings of the 33th ACM International Conference on Multimedia}{October 27--31, 2025}{Dublin, Ireland}
\acmBooktitle{Proceedings of the 33th ACM International Conference on Multimedia (MM '25), October 27--31, 2025, Dublin, Ireland}

\begin{document}

\title{CoFi-Dec: Hallucination-Resistant Decoding via Coarse-to-Fine Generative Feedback in  Large Vision-Language Models}

\author{Zongsheng Cao}
\authornote{Equal contribution}
\email{agiczsr@gmail.com}
\affiliation{%
  \institution{Researcher}
  \country{}
}

\author{Yangfan He}
\authornotemark[1]
\email{he00577@umn.edu}
\affiliation{%
  \institution{UMN 
  }
  \country{}
  }
  
\author{Anran Liu}
\authornotemark[1]
\authornote{Corresponding author}
\email{anniegogo1008@gmail.com}
\affiliation{%
  \institution{Researcher}
  \country{}
}

  \author{Jun Xie} 
  \email{xiejun@lenovo.com}
  \affiliation{%
    \institution{PCIE}
    \country{}
  }

  \author{Feng Chen}
 \email{chenfeng@lenovo.com}
 \affiliation{%
   \institution{PCIE}
   \country{}
 }

  \author{Zepeng Wang}
  \authornotemark[2]
  \email{wangzpb@lenovo.com}
  \affiliation{%
    \institution{PCIE}
    \country{}
  }

\renewcommand{\shortauthors}{Zongsheng Cao et al.}

\begin{abstract}
  Large Vision-Language Models (LVLMs) have achieved impressive progress in multi-modal understanding and generation. However, they still tend to produce hallucinated content that is inconsistent with the visual input, which limits their reliability in real-world applications. We propose \textbf{CoFi-Dec}, a training-free decoding framework that mitigates hallucinations by integrating generative self-feedback with coarse-to-fine visual conditioning. Inspired by the human visual process from global scene perception to detailed inspection, CoFi-Dec first generates two intermediate textual responses conditioned on coarse- and fine-grained views of the original image. These responses are then transformed into synthetic images using a text-to-image model, forming multi-level visual hypotheses that enrich grounding cues. 
To unify the predictions from these multiple visual conditions, we introduce a Wasserstein-based fusion mechanism that aligns their predictive distributions into a geometrically consistent decoding trajectory. This principled fusion reconciles high-level semantic consistency with fine-grained visual grounding, leading to more robust and faithful outputs. 
Extensive experiments on six hallucination-focused benchmarks show that CoFi-Dec substantially reduces both entity-level and semantic-level hallucinations, outperforming existing decoding strategies. The framework is model-agnostic, requires no additional training, and can be seamlessly applied to a wide range of LVLMs. The implementation is available at \url{https://github.com/AI-Researcher-Team/CoFi-Dec}.
\end{abstract}





\maketitle

\input{sections/1_introduction}

\input{sections/2_related_work}
\input{sections/3_method}

\input{sections/4_experiment}
\input{sections/5_conclusion}

\bibliographystyle{ACM-Reference-Format}
\balance
\bibliography{main}
\newpage
\appendix
\input{sections/6_appendix.tex}


\end{document}

%% file: sections/1_introduction.tex
\vspace{-10pt}

\section{Introduction}
\vspace{-4pt}
\label{sec:intro}

In recent years, large vision-language models (LVLMs) have attracted extensive attention, and achieved impressive results across a range of 
multimodal tasks, including image captioning and visual question answering, by extending the 
representational power of Large Language Models (LLMs) to process visual data \cite{bai2023qwen,ye2024mplug}. Despite their success in jointly modeling visual and 
textual information, LVLMs remain prone to 
generating hallucinations: outputs that contradict or 
deviate from the actual visual input. \cite{li2023evaluating,gunjal2024detecting,yin2023woodpecker,wu2024evaluating}. Such behavior poses serious risks of misinformation, undermining the trustworthiness of these models and limiting their applicability in safety-critical or real-world scenarios~\cite{liu2024survey,bai2024hallucination,zhao2024fact}.

A growing body of work attributes this issue to the models ' tendency to overﬁt to language priors, a byproduct of imbalanced training data that leads them to prioritize linguistic patterns over grounded visual evidence \cite{leng2024mitigating,bai2024hallucination,liu2024survey}. To address this, several methods have focused on hallucination suppression through additional supervision or enhanced training schemes~\cite{chen2024alleviating,li2023evaluating,zhang2024reflective}. While these approaches have shown effectiveness, their dependence on extensive retraining and additional annotated data significantly limits their scalability and usability in downstream applications.

To overcome these limitations, a newer class of approaches shifts focus from training-time interventions to decoding-time strategies~\cite{huang2024opera,deng2024seeing,kim2024code}. In particular, recent advances in contrastive decoding, achieved without any model retraining, have shown promising results in reducing hallucinations \cite{li2023contrastive}. These methods work by contrasting token predictions conditioned on faithful visual input against those generated under intentionally biased or weakened conditions, such as missing or corrupted images~\cite{favero2024multi,leng2024mitigating}, noisy prompts~\cite{wang2024mitigating2}, or truncated intermediate layers~\cite{chuang2024dola}. This inference-time strategy provides an efficient and generalizable pathway to mitigate hallucinations without incurring the overhead of additional training. Moreover, some prior work has explored hallucination mitigation through either architectural modifications~\cite{li2023blip,liu2023visual,tong2024eyes}, fine-tuning with annotated data, or post-hoc filtering. More recently, decoding-time strategies have attracted increasing attention due to their training-free and model-agnostic properties. Among these, feedback-based decoding methods attempt to revise or verify model predictions by incorporating auxiliary signals, such as retrieved factual knowledge or images synthesized from textual hypotheses. 

\begin{figure}[t]
  \begin{center}
   \includegraphics[width=0.48\textwidth]{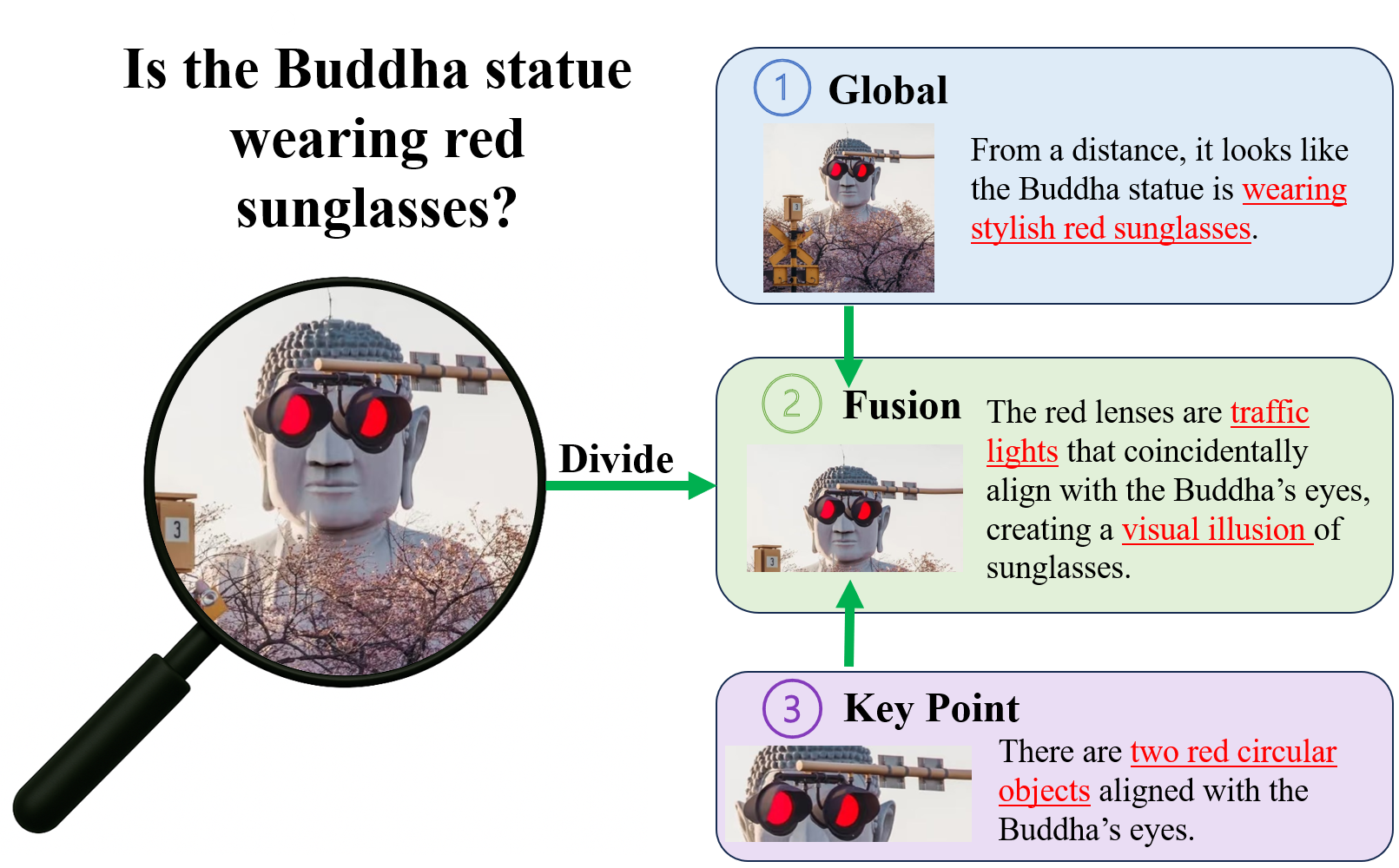}
  \end{center}
  \caption{The illustration of the importance of combining coarse-grained and fine-grained visual information to avoid misleading global interpretations. By decomposing the scene, we uncover that the perceived red sunglasses are actually traffic lights, highlighting the need for multi-scale reasoning in visual understanding.}
  \label{fig:intro}
\end{figure}

While conceptually promising, these approaches are often limited in three critical aspects. As shown in Fig.\ref{fig:intro}.
First, most feedback-based methods rely solely on the original image input and operate at a single resolution, overlooking the inherent multiscale structure of visual information. As a result, they may fail to detect inconsistencies that manifest at different semantic levels, such as global layout versus object details. Second, generative feedback is typically used in a post-hoc fashion, serving merely as a reference for re-ranking or answer replacement. This weakens its influence on the actual token-level generation process. Third, existing strategies rarely incorporate feedback in a fine-grained, step-by-step manner that allows cumulative correction and refinement during decoding.

In this work, we propose \textbf{CoFi-Dec}, a new decoding framework that addresses these limitations through \textit{coarse-to-fine guided generative feedback}. Our key insight stems from the observation that hallucinations often result from imbalanced attention across visual granularities. To address this, we design a human-inspired decoding process that mimics perceptual strategies in visual cognition, where coarse scanning precedes focused inspection. Motivated by previous work \cite{zhang2025self,liu2025mmbench}, given an input image, CoFi-Dec first constructs two visual pathways by decomposing the image into coarse-grained and fine-grained representations. The coarse view captures global context by uniformly downsampling the image into low-resolution patches, while the fine view highlights local semantics through high-resolution crops around salient or ambiguous regions. These two complementary inputs are used to generate independent textual responses using an LVLM. Each response serves as a semantic hypothesis derived from a different perceptual lens.
To convert these hypotheses into actionable feedback, we employ a generative vision model (e.g., Stable Diffusion) to synthesize two pseudo-images corresponding to the coarse- and fine-grained responses. These generated images serve as the model's self-imagined reflections of its own beliefs, capturing what the model “sees” at different levels of abstraction. Rather than treating these synthetic images as auxiliary evidence, we integrate them along with the original image into the decoding loop. This integration guides subsequent token generation.

At each decoding step, CoFi-Dec computes three conditional token distributions, each based on a distinct visual condition: the original image, the synthesized coarse image, and the synthesized fine image. These distributions are then fused using a \textit{Wasserstein barycenter optimization}, which computes a geometry-aware consensus across the semantic token space. Unlike naïve averaging, this method respects the underlying structure of the vocabulary space and encourages smooth, semantically meaningful corrections.

This multi-path decoding design allows CoFi-Dec to cross-reference visual semantics at multiple scales, verify generated content from self-imagined feedback, and continuously refine output in a self-correcting fashion. Importantly, the entire framework is training-free, modular, and compatible with a wide range of pre-trained LVLMs, making it practical for real-world deployment.

To summarize, our main contributions are:
\begin{itemize}
    \item We propose \textbf{CoFi-Dec}, a novel training-free decoding framework that incorporates coarse-to-fine generative visual feedback to mitigate hallucinations in LVLMs.
    \item We design a multi-granular visual decomposition mechanism that mirrors human-like inspection patterns, enabling distinct semantic views for generative self-verification.
    \item We introduce a Wasserstein-based fusion strategy that jointly considers original and feedback-conditioned predictions to produce geometrically consistent token-level decoding.
    \item We conduct extensive experiments across six hallucination-sensitive benchmarks, demonstrating that CoFi-Dec consistently outperforms state-of-the-art decoding baselines in hallucination mitigation.
\end{itemize}

%% file: sections/2_related_work.tex
\section{Related Work}

\textbf{Hallucination in LVLMs}.  
Autoregressive large language models such as LLaMA 2 ~\citep{touvron2023llama}, PaLM 2 ~\citep{chowdhery2023palm}, and Vicuna ~\citep{chiang2023vicuna} have catalysed a rapid shift from text-only modelling to large vision–language models (LVLMs) ~\citep{liu2023visual,dai2024instructblip,bai2023qwen,ye2024mplug}. In a typical LVLM, raw image features are passed through a lightweight adapter or projection layer, allowing pixel-level signals to share a common embedding space with tokens; this simple alignment lets a single backbone reason fluidly across modalities and has driven state-of-the-art results on image captioning, visual question answering, and allied benchmarks ~\citep{liu2024survey,bai2024hallucination}.

Performance gains, however, mask a persistent liability, which is called hallucination. LVLMs still invent objects, attributes, or relations that are absent from the scene ~\citep{li2023evaluating,liu2024survey,bai2024hallucination}. Current mitigation strategies fall into three broad camps. Alignment methods (e.g.\ RLHF) try to steer generation toward fidelity ~\citep{gunjal2024detecting,sun2023aligning}. Training-time regularisers embed auxiliary losses that penalise visual-textual mismatch ~\citep{jiang2024hallucination,chen2023mitigating}. Data-centric approaches toughen the model with noise or adversarial samples ~\citep{yue-etal-2024-less,liu2023mitigating}. A complementary, post-hoc line edits outputs via verifier–editor cascades ~\citep{zhou2024analyzing,yin2023woodpecker}. While all show promise, they typically demand heavy curation or repeated fine-tuning, which hampers scalability and real-world adoption.

A growing body of work now targets training-free, zero-tuning add-ons that enhance pretrained LVLMs without touching their weights. Two main families have emerged. Contrastive-decoding schemes~\citep{leng2024mitigating,favero2024multi} suppress hallucinations by ranking or subtracting continuations produced under different sampling rules, whereas guided-decoding techniques~\citep{chen2024halc,deng2024seeing,woo2024ritual} inject auxiliary signals directly into the token-generation loop. Most recently, \citet{zhang2025self} extends this trend: they create synthetic images as iterative “visual critiques,” allowing the model to refine its own drafts and curb hallucinations in a manner reminiscent of human double-checking.

\noindent\textbf{Text-to-Image Synthesis}.Text-to-image synthesis seeks to translate a free-form sentence into a picture that is both semantically faithful and visually convincing \citep{zhu2019dm,ge2023expressive}. Breakthroughs in deep generative modelling \citep{goodfellow2014generative,zhan2023multimodal} have propelled the field forward, producing three flagship families: diffusion models \citep{ho2020denoising,karras2022elucidating,nichol2022glide,saharia2022photorealistic,rombach2021highresolution}, generative adversarial networks (GANs) \citep{sauer2023stylegan,kang2023scaling}, and autoregressive token predictors \citep{yu2022scaling,chang2023muse}. Diffusion methods now dominate thanks to their ability to render highly detailed, photorealistic imagery while offering precise user control \citep{yang2023diffusion,croitoru2023diffusion}. When exposed to web-scale corpora such as LAION \citep{schuhmann2022laion}, they learn tight text–image correspondences that translate into strong zero-shot performance on downstream tasks, from image classification \citep{li2023your} to semantic segmentation \citep{amit2021segdiff,wolleb2022diffusion}.

Recent work by \citet{jiao2024img} shows that diffusion generators can improve fine-grained recognition. Using Stable Diffusion XL \citep{podell2024sdxl}, they built the Img-Diff corpus, which contains paired images with subtle variations, and demonstrated that fine-tuning LVLMs on this synthetic set increases accuracy on multiple VQA benchmarks. In contrast to such data-augmentation-and-retraining pipelines, \citet{zhang2025self} takes a training-free approach. During inference, a pretrained diffusion model acts as a visual feedback loop: it converts the LVLM’s initial text into synthetic images and feeds them back to the model for self-correction. This iterative process improves factual consistency and visual grounding while keeping the original LVLM weights unchanged, thus maintaining its deployment footprint. However, figures often contain both coarse-grained and fine-grained features, which previous work has overlooked. Addressing these features remains an open challenge for hallucination resistance.

%% file: sections/3_method.tex
\vspace{-3pt}
\section{Methodology}
\begin{figure*}[t]
  \begin{center}
     \makebox[\textwidth]{\includegraphics[width=\textwidth]{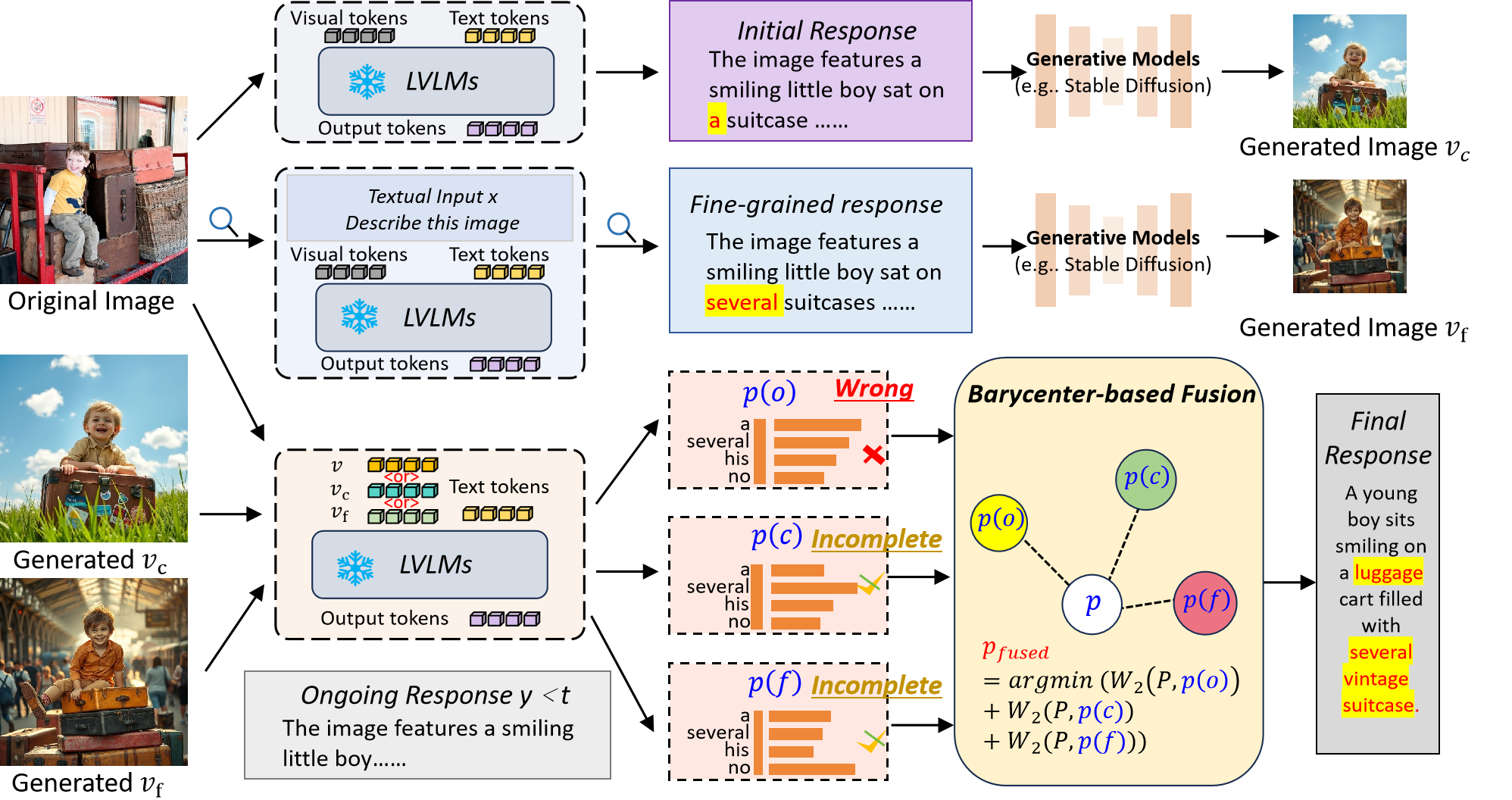}}
  \end{center}
  \vspace{-10pt}
  \caption{\textbf{Overview of our proposed CoFi-Dec}. Our method follows a coarse-to-fine generative feedback framework to enhance the fidelity of image-grounded text generation. By generating both coarse- and fine-grained textual descriptions from the original image and synthesizing corresponding images, we obtain multi-perspective visual feedback. These signals are then fused using a Wasserstein barycenter to produce a final response that is semantically rich, visually grounded, and resistant to hallucination.} 
  \label{fig:overview}
\end{figure*}

\vspace{-2pt}
\label{sec:method}
In this paper, we introduce \textbf{CoFi-Dec}, a novel training-free framework designed to enhance the reliability of LVLM responses through recursive refinement using feedback from a text-to-image generative model, as depicted in Figure~\ref{fig:overview}.

\noindent\textbf{Problem Setting.}  
We assume access to a vision-language model (LVLM) parameterized by $\theta$, which receives a visual input $v$ and a textual query $\mathbf{x}$, and aims to generate a coherent and relevant textual response sequence $\mathbf{y}$ in an autoregressive fashion. The image $v$ is first encoded by a vision encoder and subsequently mapped into a sequence of visual tokens via a vision-language projection module, such as a Q-Former~\citep{li2023blip} or a linear projection layer~\citep{liu2023visual}, which aligns the visual features with the language model's embedding space. These visual tokens, together with the tokenized textual query, are fed into the language encoder to condition the generative process.
Formally, the generation at each time step $t$ is governed by the following distribution:
\begin{equation}
y_t \sim p_{\theta}(y_t | v, \mathbf{x}, \mathbf{y}_{<t}) \propto \exp f_{\theta}(y_t | v, \mathbf{x}, \mathbf{y}_{<t}),
\end{equation}
where $y_t$ denotes the token generated at step $t$, and $\mathbf{y}_{<t} = [y_0, \dots, y_{t-1}]$ represents the sequence of previously generated tokens. The function $f_{\theta}$ outputs the unnormalized logit scores over the vocabulary $\mathcal{V}$, which are then transformed into probabilities for sampling. This autoregressive decoding continues until an end-of-sequence token is generated, yielding the final response $\mathbf{y} = [y_0, \dots, y_T]$.

     \subsection{Generative Feedback with Multi-Granular Conditioning}

     Despite recent advances, hallucination remains a persistent challenge for Large Vision-Language Models (LVLMs), especially in scenarios requiring precise grounding of visual entities. Existing decoding strategies, leverage auxiliary synthetic visual signals to refine the output, yet they predominantly rely on the original single-scale image input. Such representations may fail to capture critical visual nuances that are essential for resolving ambiguity or verifying semantic consistency.
     
     Motivated by the multi-scale processing mechanism in human vision, where observers ﬁrst perform a coarse scan of the visual ﬁeld before engaging in ﬁne-detail scrutiny, we propose a novel decoding framework that jointly incorporates coarse-grained, ﬁne-grained, and original visual cues. By structuring the visual input into a hierarchical multi-resolution context and integrating it into a generative feedback loop, our model achieves higher robustness and fidelity in grounding visual content during generation.
     
     \noindent\textbf{Hierarchical Visual Decomposition.}
     Given an input image $I_0 \in \mathbb{R}^{H \times W \times 3}$ and an associated textual prompt $T$, we first decompose the image into two complementary sets of patches that capture different granularity levels:
     
     \begin{itemize}
         \item \textbf{Coarse-grained views} $\mathcal{I}_c = \{I_c^1, \ldots, I_c^n\}$: Each $I_c^i$ is obtained by uniformly dividing $I_0$ into $n$ non-overlapping low-resolution patches, preserving the global spatial structure while discarding fine details. This resembles a downsampled global scan that preserves contextual semantics.
         \item \textbf{Fine-grained views} $\mathcal{I}_f = \{I_f^1, \ldots, I_f^m\}$: Each $I_f^j$ is a high-resolution crop focusing on salient or uncertain regions, derived from either learned attention maps or region proposal algorithms. These patches highlight local discriminative features that are potentially omitted by coarse processing.
     \end{itemize}
     In this way,
     the full multi-resolution visual input is denoted by the unified set:
     \begin{equation}
      \mathcal{I} = I_0 \cup \mathcal{I}_c \cup \mathcal{I}_f.
     \end{equation}

     To this end, the initial response $R_0$ is generated by an LVLM conditioned only on the original image $I_0$ and the prompt $T$:
     \begin{equation}
     R_0 = \text{LVLM}(I_0, T).
     \end{equation}
 We then perform two conditional generations under different granularity contexts from arse- and fine-grained perspectives:
     \begin{align}
     R_c &= \text{LVLM}( \mathcal{I}_c , T), \\
     R_f &= \text{LVLM}( \mathcal{I}_f , T).
     \end{align}
     To enable self-verification, a generative vision model $G$ (e.g., Stable Diffusion) is used to synthesize a pseudo-image based on the textual hypothesis.
The synthesized image serves as a reflection of the model's internal belief and facilitates visual grounding in subsequent refinement stages. We obtain the corresponding synthesized figures $v_c$ and $v_f$ as follows:
  \begin{equation}
v_c=G(R_c), v_f=G(R_f).
     \end{equation}

\subsection{Self-Correcting Decoding with Generative Feedback}
\label{sec:token}

In this part, we describe how to incorporate the complementary cues from coarse-grained, fine-grained, and original image views into the decoding process in a more fine-grained and dynamic manner. While previous sections focus on generating separate responses conditioned on different visual granularities, these responses are inherently static and limited in their ability to reconcile semantic discrepancies across views. To address this, we move beyond fixed-level conditioning and explore a decoding-time integration strategy that operates at the token level. Our goal is to enable more precise alignment between multimodal inputs and language outputs by leveraging 
the unique strengths of each visual perspective: the original image captures layout and raw semantics, the 
coarse-grained view emphasizes holistic structure, and the ﬁne-grained view provides object-level precision.

Specifically, rather than relying on isolated end-to-end responses from each visual condition, we propose a \emph{self-correcting decoding} strategy that dynamically fuses the predictive signals across granularities for each generated token. This allows the model to adaptively weigh global context, structural composition, and localized evidence in a principled way at every step of generation, thus enabling more faithful and grounded outputs.

Let $v$, $v_c$, and $v_f$ denote the visual embeddings extracted from the original image, the synthesized coarse-grained context, and the fine-grained visual reference, respectively. For a given prompt $\mathbf{x}$ and previously generated sequence $\mathbf{y}_{<t}$, the model produces three conditional probability distributions over the vocabulary for the next token $y_t$:
\begin{equation}
  \begin{aligned}
p_{\theta}(y_t | v, \mathbf{x}, \mathbf{y}_{<t})\!&=\!\mathtt{Softmax}\!\left[f_\theta(y_{t}| v,\mathbf{x},\mathbf{y}_{<t})\right]\!,\\
p_{\theta}(y_t | v_c, \mathbf{x}, \mathbf{y}_{<t})\!&=\!\mathtt{Softmax}\!\left[f_\theta(y_{t}| v_c,\mathbf{x},\mathbf{y}_{<t})\right],  \\
p_{\theta}(y_t | v_f, \mathbf{x}, \mathbf{y}_{<t})\!&=\!\mathtt{Softmax}\!\left[f_\theta(y_{t}| v_f,\mathbf{x},\mathbf{y}_{<t})\right].  
  \end{aligned}
\end{equation}
To obtain a unified prediction distribution that effectively integrates multi-granular visual information, we leverage the concept of Wasserstein barycenters to fuse the token-level output distributions conditioned on the original image $v$, the coarse-grained reference $v_c$, and the fine-grained reference $v_f$. Unlike simple averaging or heuristic weighting schemes, the Wasserstein barycenter provides a principled way to compute a central distribution that minimizes the overall transportation cost with respect to the constituent distributions, thereby preserving the underlying semantic geometry of the token space.

Formally, for a fixed decoding timestep $t$, we denote the three predictive distributions over the vocabulary as $P^{(v)}_t = p_\theta(y_t | v, \mathbf{x}, \mathbf{y}_{<t})$, $P^{(c)}_t = p_\theta(y_t | v_c, \mathbf{x}, \mathbf{y}_{<t})$, and $P^{(f)}_t = p_\theta(y_t | v_f, \mathbf{x}, \mathbf{y}_{<t})$, each represented as a probability vector in the $|\mathcal{V}|$-dimensional simplex, where $\mathcal{V}$ is the vocabulary.

To compute the fused distribution $P^{(\text{fused})}_t$ that serves as the final prediction for $y_t$, we solve the following optimization problem:

\begin{equation}
P^{(\text{fused})}_t = \arg\min_{P \in \Delta^{|\mathcal{V}|}} \left( W(P, P^{(v)}_t) +   W(P, P^{(c)}_t) +  W(P, P^{(f)}_t) \right),
\label{eq:wasserstein-barycenter}
\end{equation}
where $W(P, Q)$ denotes the Wasserstein distance between two distributions $P$ and $Q$. This formulation ensures that the fused distribution is not only probabilistically valid but also geometrically faithful to the structure of the input distributions.

Once the barycenter $P^{(\text{fused})}_t$ is computed for each timestep $t$, it is used in place of the individual conditional distributions for token selection during generation. Notably, this fusion mechanism allows the model to adaptively reconcile the high-level contextual cues from $v_c$ with the fine-grained semantic alignment from $v_f$, grounded by the original image $v$, yielding a robust and geometrically consistent prediction. This Wasserstein-based fusion strategy thus constitutes a key component of our model's capability to perform multi-granular visual reasoning during language generation.

\begin{remark}
  Our proposed framework introduces a unified decoding paradigm that explicitly models multi-resolution visual reasoning and aligns it with generative feedback. Compared to prior work that treats feedback as a post-hoc correction, we integrate it into a multi-path decoding framework where each visual granularity serves as an independent yet complementary source of evidence.
  By mirroring human-like inspection patterns, in which a global preview is followed by 
  selective zoom-in, the model acquires a more holistic and reliable understanding of the visual 
  scene. This reduces reliance on spurious correlations and encourages grounded generation, especially in scenarios with ambiguous, cluttered, or fine-detailed imagery.
\end{remark}

%% file: sections/4_experiment.tex

\section{Experiments}
\label{sec:experiment}

In this section, we evaluate our model through a suite of benchmark trials designed to gauge its ability to suppress hallucinations in LVLMs, and we juxtapose the resulting metrics with those reported by the strongest contemporary baselines.
\input{tabs/pope}
\subsection{Experimental Settings}

\noindent\textbf{LVLMs}.  
To evaluate our model, we ran a comprehensive test suite on three headline open-source LVLMs: the upgraded \textbf{LLaVA-1.5}~\citep{liu2024improved}, \textbf{InstructBLIP}~\citep{dai2024instructblip}, and Alibaba’s \textbf{Qwen-VL}~\citep{bai2023qwen}.  The first two share an identical textual backbone, \textit{Vicuna-7B}~\citep{chiang2023vicuna}, itself a dialogue-oriented adaptation of \textit{LLaMA}~\citep{touvron2023llama}.  By contrast, Qwen-VL is anchored in the 7-billion-parameter Qwen family.  For our own trials, we simply plug the proposed CoFi-Dec module into the publicly released \textit{Qwen-VL-Chat} checkpoint, leaving all original weights intact.

\noindent\textbf{Benchmarks}.  Following previous work \cite{huang2024opera,zhang2025self}, 
our empirical study taps into six publicly available testbeds that jointly probe a model's resistance to hallucination and its overall visual–reasoning prowess:
(1) \textbf{POPE}~\citep{li2023evaluating} focuses on object hallucination by presenting binary questions that test whether the model can correctly identify the presence or absence of objects in an image.  
(2) \textbf{CHAIR}~\citep{rohrbach2018object} examines hallucinations in free-form image captions, requiring models to describe images randomly sampled from the MSCOCO validation set, with attention to visual grounding.  
(3) \textbf{MME- Hallucination} \citep{fu2023mme} provides a fine-grained evaluation of both object-level and attribute-level hallucinations via four sub-tasks: \textit{existence}, \textit{count}, \textit{position}, and \textit{color}.  
(4) \textbf{MMBench}~\citep{liu2025mmbench} serves as a broad-spectrum evaluation suite, covering 20 aspects of multi-modal reasoning to test LVLMs’ general understanding capabilities.  
(5) \textbf{MMVP}~\citep{tong2024eyes} targets fine-grained recognition through CLIP-blind image pairs, challenging models with binary questions across 150 curated examples.  
(6) \textbf{LLaVA-Bench} comprises 24 images ranging from complex real-world scenes to artistic renderings (e.g., memes, paintings, and sketches), paired with 60 intricate questions designed to test both visual perception and contextual comprehension.

\begin{table*}[t]
      \begin{center}
      \begin{small}
      \setlength{\tabcolsep}{5pt} 
      \caption{\textbf{Results on CHAIR~\citep{rohrbach2018object} benchmark.} We limit the maximum number of new tokens to 64. Lower ($\downarrow$) CHAIR$_S$, CHAIR$_I$ and higher ($\uparrow$) recall and length indicate better performance. The best results in each setting are \textbf{bolded}, and the second-best are \underline{underlined}.}
      \label{tab:CHAIR}
      \resizebox{\textwidth}{!}{
          \begin{tabular}{lcccccccc}
              \toprule
                \multirow{2}{*}[-0.5ex]{\textbf{Method}}  &  \multicolumn{4}{c}{\textbf{LLaVA-1.5}} & \multicolumn{4}{c}{\textbf{InstructBLIP}} \\
               \cmidrule(lr){2-5}\cmidrule(lr){6-9}
                   & CHAIR$_S$ $\downarrow$ & CHAIR$_I$ $\downarrow$ & Recall $\uparrow$ & Length $\uparrow$   & CHAIR$_S$ $\downarrow$ & CHAIR$_I$ $\downarrow$ & Recall $\uparrow$ & Length $\uparrow$ \\
               \midrule
               Regular & 26.1 & 9.3 & 58.6 & 53.3 &  31.3 & 11.0  & 59.1 & 53.5\\
                VCD & 24.3 & 7.8 & 63.4 & \underline{54.4} &  29.9 & 10.2  & 61.9 & 54.1\\
                M3ID & 21.5 & \textbf{\underline{6.1}}& 64.1 & 53.6 & 30.7 & 10.3 & 62.5 & 53.3\\
                RITUAL & 22.3 & 6.8  & 63.1 & 54.7 &  26.7 & 8.8 & 63.5 & \underline{55.2}\\
                \RebuttalRevision{Woodpecker} & 24.8 & 7.6  & 60.9 & 49.6 &  31.1 & 10.7 & 62.4 & 51.4\\
                \RebuttalRevision{HALC} & 21.6 & 7.2  & \underline{64.9} & 53.5 &  24.6 & 8.1 & \underline{64.2} & 54.8\\
                DeFG& \underline{18.4}& 6.4& 62.9& 54.3& \underline{24.0}& \underline{7.7}& 63.5& 55.0\\
               \cc \textbf{Ours}& \cc \textbf{18.1}& \cc \textbf{6.1}& \cc \textbf{65.2}& \cc \textbf{55.7}& \cc \textbf{23.2}& \cc \textbf{7.3}& \cc \textbf{68.9}& \cc \textbf{55.8}\\
               \bottomrule
              \end{tabular}
      }
      \end{small}
      \end{center}
      \vspace{-10pt}
\end{table*}

\begin{table*}[t!]
      \begin{center}
      \begin{small}
      \setlength{\tabcolsep}{6pt} 
      \caption{Results on MME-Hallucination~\citep{fu2023mme} and MMBench~\citep{liu2025mmbench} benchmark. We present the average MME scores along with their standard deviations over three random seeds for each subset. Additionally, the final column reports the overall accuracy of each method on the MMBench benchmark. Higher values ($\uparrow$) denote better performance. The best results are highlighted in \textbf{bold}, while the second-best are marked with \underline{underlining}.}
      \label{tab:MME}
      \resizebox{\textwidth}{!}{
          \begin{tabular}{lcccccc}
              \toprule
               \multirow{2}{*}[-0.5ex]{\textbf{Method}}  &  \multicolumn{2}{c}{\textbf{Object-level}} & \multicolumn{2}{c}{\textbf{Attribute-level}} & \multirow{2}{*}[-0.5ex]{\textbf{MME Score $\uparrow$}} & \multirow{2}{*}[-0.5ex] {\RebuttalRevision{\textbf{MMBench $\uparrow$}}}\\
               \cmidrule(lr){2-3}\cmidrule(lr){4-5}
                  & Existence $\uparrow$ & Count $\uparrow$ & Position $\uparrow$ & Color $\uparrow$   & \\
               \midrule
               Regular &  173.75 {\tiny ($\pm4.79$)} & 121.67 {\tiny ($\pm12.47$)} & 117.92 {\tiny ($\pm3.69$)\phantom{0}} & 149.17 {\tiny ($\pm7.51$)\phantom{0}} &  562.50 {\tiny ($\pm3.96$)\phantom{0}} & \RebuttalRevision{64.1} \\
                DoLa  & 176.67 {\tiny ($\pm2.89$)} & 113.33 {\tiny ($\pm10.41$)} & 90.55 {\tiny ($\pm8.22$)} & 141.67 {\tiny ($\pm7.64$)\phantom{0}} &  522.22 {\tiny ($\pm16.78$)} & \RebuttalRevision{63.8} \\
                OPERA  & 183.33 {\tiny ($\pm6.45$)} & 137.22 {\tiny ($\pm6.31$)\phantom{0}} & 122.78 {\tiny ($\pm2.55$)\phantom{0}} &155.00 {\tiny ($\pm5.00$)\phantom{0}}  &  598.33 {\tiny ($\pm10.41$)} & \RebuttalRevision{64.4} \\
                VCD  & 186.67 {\tiny ($\pm5.77$)} & 125.56 {\tiny ($\pm3.47$)\phantom{0}} & 128.89 {\tiny ($\pm6.73$)\phantom{0}} & 139.45 {\tiny ($\pm12.51$)} &  580.56 {\tiny ($\pm15.13$)} & \RebuttalRevision{64.5} \\
                M3ID  & 186.67 {\tiny ($\pm5.77$)} &  128.33 {\tiny ($\pm10.41$)} & \underline{131.63} {\tiny ($\pm5.00$)\phantom{0}} & 151.67 {\tiny ($\pm20.88$)} &  597.50 {\tiny ($\pm20.35$)}  & \RebuttalRevision{64.3} \\
               RITUAL  & \underline{187.53} {\tiny ($\pm2.89$)} & 139.58 {\tiny ($\pm7.64$)\phantom{0}}  & 125.00 {\tiny ($\pm10.27$)} & \underline{163.33} {\tiny ($\pm6.87$)\phantom{0}} & \underline{626.29} {\tiny ($\pm20.38$)} & \RebuttalRevision{64.0}    \\
              \RebuttalRevision{Woodpecker} & \RebuttalRevision{\underline{187.53} {\tiny ($\pm2.89$)}} & \RebuttalRevision{126.25 {\tiny ($\pm2.17$)\phantom{0}}} & \RebuttalRevision{126.66 {\tiny ($\pm2.89$)\phantom{0}}} & \RebuttalRevision{149.17 {\tiny ($\pm17.34$)}} &  \RebuttalRevision{589.58 {\tiny ($\pm10.00$)}} & \RebuttalRevision{64.2}\\
              \RebuttalRevision{HALC} & \RebuttalRevision{183.33 {\tiny ($\pm0.00$)}} & \RebuttalRevision{133.33 {\tiny ($\pm5.77$)\phantom{0}}} & \RebuttalRevision{109.58 {\tiny ($\pm3.69$)\phantom{0}}} &\RebuttalRevision{155.00 {\tiny ($\pm5.00$)\phantom{0}}}  &  \RebuttalRevision{581.24 {\tiny ($\pm9.07$)\phantom{0}}} & \RebuttalRevision{64.4}\\
              \RebuttalRevision{DeFG} & \RebuttalRevision{188.33 {\tiny ($\pm2.89$)}} & \RebuttalRevision{\underline{142.50} {\tiny ($\pm6.64$)\phantom{0}}} & \RebuttalRevision{131.33 {\tiny ($\pm3.85$)\phantom{0}}} &\RebuttalRevision{163.17 {\tiny ($\pm3.47$)\phantom{0}}}  &  \RebuttalRevision{625.83 {\tiny ($\pm9.18$)\phantom{0}}} & \RebuttalRevision{\underline{65.2}}
              
              \\
               
               \cc \textbf{Ours} & \cc \textbf{190.26} {\tiny ($\pm2.31$)} & \cc \textbf{144.43} {\tiny ($\pm5.27$)\phantom{0}}  & \cc \textbf{133.71} {\tiny ($\pm3.14$)\phantom{0}} & \cc \textbf{165.62} {\tiny ($\pm4.08$)\phantom{0}} & \cc \textbf{627.39} {\tiny ($\pm8.53$)\phantom{0}} & \cc \RebuttalRevision{\textbf{65.9}} \\
              \bottomrule
          \end{tabular}
      }
      \vspace{-15pt}
      \end{small}
      \end{center}
\end{table*}
\noindent\textbf{Baselines}.  
We begin with a straightforward baseline in which decoding is performed in the conventional manner, 
that is, each token is sampled directly from the softmax-normalized output probabilities. Beyond this, we evaluate our approach against three state-of-the-art decoding strategies: VCD~\citep{leng2024mitigating}, M3ID~\citep{favero2024multi}, and RITUAL~\citep{woo2024ritual}. For benchmarks such as CHAIR~\citep{rohrbach2018object} and MME-Hallucination~\citep{fu2023mme}, we expand the comparison set to include additional recent methods, namely Woodpecker~\citep{chen2024halc}, HALC~\citep{chen2024halc}, DoLa~\citep{chuang2024dola}, OPERA~\citep{huang2024opera}, and DeGF~\citep{zhang2025self}. Performance results for these baselines are derived from our own re-implementation based on their officially released source code.

\noindent\textbf{Implementation Details}.  
In all experiments, we maintain consistency with the standard input formatting adopted by LLaVA-1.5~\citep{liu2024improved} and InstructBLIP~\citep{dai2024instructblip}. 
To account for variability, we run each MME benchmark experiment three times with different random seeds and report both the mean accuracy and standard deviation.

\subsection{Results and Discussions}
\noindent\textbf{Results on POPE}.  
Table~\ref{tab:POPE} presents a comparative analysis of our proposed method and several strong baselines on the POPE benchmark, evaluated under three distinct negative sampling strategies across three different datasets.  More details can refer to Appendix. 
Our approach consistently yields superior performance over all competitors across the 18 evaluation settings, achieving the highest accuracy, precision, and F1 score on both LVLM architectures. Specifically, we observe significant gains in different metrics when compared to the next-best method. These results indicate that integrating a generative reference allows the model to better capture fine-grained visual cues, effectively mitigating object hallucinations. Additionally, while many decoding approaches exhibit overconﬁdence and often default to afﬁrmative answers, our self-reﬁning decoding strategy demonstrates a more cautious response pattern. This is reﬂected in its consistently higher precision, indicating a stronger ability to avoid false positives and resist generating misleading outputs.

An important observation is the method’s robustness in the more difficult popular and adversarial scenarios. Unlike the random setting, these configurations are characterized by frequent inclusion and co-occurrence of non-existent negative objects~\citep{li2023evaluating}, which tend to trigger hallucinations in existing models. Despite this increased complexity, our method experiences significantly smaller drops in performance compared to all other baselines. This suggests that our generative feedback mechanism not only enhances the model's visual grounding but also helps it discern misleading object associations, thus improving its resistance to context-driven hallucinations.

\noindent\textbf{Results on CHAIR}.  
We evaluate our approach on the open-ended image captioning task and compare it with several state-of-the-art decoding strategies. The CHAIR scores, recall values, and average response lengths are summarized in Table~\ref{tab:CHAIR}. These evaluations are conducted on two representative LVLMs and consistently show that our method outperforms all baselines. Notably, our approach surpasses the second-best method by margins of significant gain on the CHAIR$_S$ metric, respectively. Furthermore, it generates more informative responses than standard decoding, as reflected by its higher recall and longer average output length.
These findings highlight the strength of our generative feedback mechanism in enhancing the fidelity and richness of model outputs. By leveraging synthesized visual references during decoding, our method helps the model better ground its predictions in visual content, leading to a significant reduction in hallucinated objects during caption generation.

\begin{figure*}[t]
  \centering
  \includegraphics[width=0.95\linewidth]{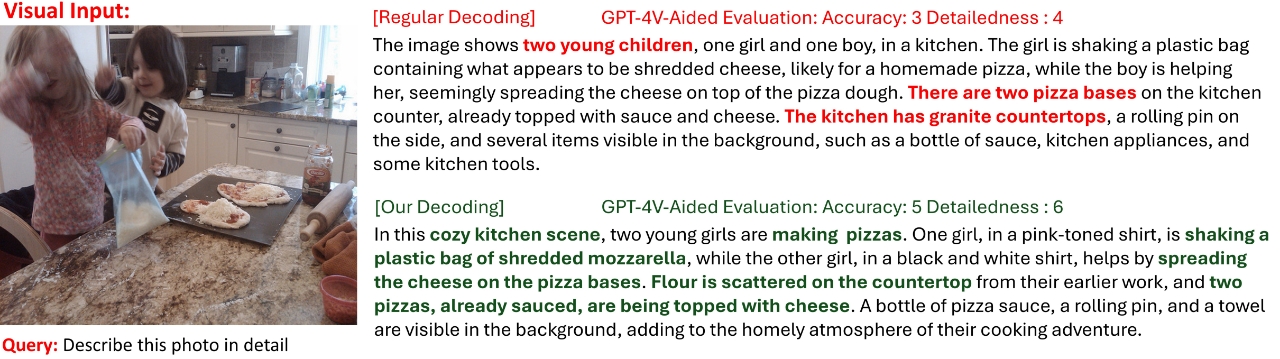}
  \caption{Case study on the LLaVA-Bench benchmark. Responses from standard decoding and our method (LLaVA-1.5) are shown with GPT-4V-assisted evaluations. Hallucinated and correct contents are highlighted in \textcolor{darkred}{red} and \textcolor{darkgreen}{green}, respectively.} 
  \label{fig:llavabench}
  \end{figure*}

  \noindent\textbf{Results on MME-Hallucination and MMBench}.  
To evaluate our method beyond object-level hallucinations, we further conduct experiments on the MME-Hallucination benchmark, which covers both object existence and attribute-based hallucination scenarios. As reported in Table~\ref{tab:MME}, our approach consistently surpasses competing baselines, achieving notable improvements in the total score metric, outperforming the second-best method by +18.19 on LLaVA-1.5 and +21.11 on InstructBLIP. Across the eight subcategories, our method secures the highest performance in six, underscoring its robustness across different hallucination types. Notably, our approach yields significant gains on the color subset, which is particularly difficult due to its reliance on accurate interpretation of subtle visual attributes. These findings confirm the strength of our generative feedback mechanism in mitigating both coarse-grained and fine-grained hallucinations. 

More details for experiments on other datasets and more case studies can refer to the Appendix.

\begin{table}[t]
  \setlength\tabcolsep{3pt}
  \caption{Ablation study. We present the performance of our approach on multiple components.}
      \label{tab:ablation study}
      \resizebox{0.48\textwidth}{!}{
      \begin{tabular}{lcccc}
          \toprule
          Models & POPE Acc. & CHAIR$_S$ & CHAIR$_I$ & MME Score \\ 
          \midrule
          Coarse Response(sdv1.5) & 89.04 & 20.3 & 7.2 & 626.6 \\
          Fine-grained Response(sdv1.5) & 89.67 & 19.8 & 6.9 & 634.15 \\
          Fine-grained - Coarse Response(sdv1.5) & \textbf{89.78} & \textbf{18.7} & \textbf{6.3} & \textbf{644.37} \\
          Remove Wasserstein Fusion(sdv1.5) & 88.01 & 22.4 & 6.2 & 624.29 \\
          Coarse Response(sdxl-v1.0) & 88.26 & 19.9 & 6.7 & 643.62 \\
          Fine-grained Response(sdxl-v1.0) & 88.45 & 19.4 & 6.4 & 648.73 \\
          Fine-grained - Coarse Response(sdxl-v1.0) & \textbf{88.60} & \textbf{18.3} & \textbf{5.8} & \textbf{649.38} \\
          Remove Wasserstein Fusion(sdxl-v1.0) & 87.13 & 21.8 & 6.6 & 638.02 \\
          Coarse Response(sd-v2.1) & 88.36 & 20.2 & 7.3 & 634.4 \\
          Fine-grained Response(sd-v2.1) & 88.54 & 19.25 & 7.1 & 644.89 \\
          Fine-grained - Coarse Response(sd-v2.1) & \textbf{88.69} & \textbf{18.44} & \textbf{6.9} & \textbf{647.24} \\
          Remove Wasserstein Fusion(sd-v2.1) & 87.19 & 21.3 & 7.8 & 631.26 \\
          \bottomrule
      \end{tabular}
      }
  \end{table}
\subsection{Ablation Studies}
\noindent\textbf{Study for Components.}
We conduct an ablation study to evaluate the contributions of coarse-to-fine reasoning and Wasserstein fusion across different diffusion backbones (sdv1.5, sdxl-v1.0, sd-v2.1). As shown in Table \ref{tab:ablation study}, results consistently show that combining fine- and coarse-grained responses outperforms using either alone, indicating that multi-level semantic aggregation improves factual grounding and descriptive accuracy. For example, the Fine-grained -Coarse Response setting yields the best CHAIR and MME scores across all backbones, demonstrating reduced hallucinations and enhanced alignment with image content. Removing the Wasserstein fusion module leads to noticeable performance drops, confirming its key role in consolidating diverse semantic cues into a unified and accurate response. The improvements are particularly evident in the CHAIR metrics, highlighting the framework's robustness in mitigating hallucinated object mentions. Overall, these results validate the effectiveness and generalizability of our proposed fine-to-coarse feedback and fusion strategy.

\noindent\textbf{Effects of Different Generative Models}.  
Table A.3 compares several CoFi-Dec variants that utilize different versions of Stable Diffusion as the generative module, all evaluated using the same LLaVA-1.5 backbone. The results demonstrate that the performance of our method remains stable across different diffusion model choices. Regardless of the specific generative variant, all configurations deliver consistent improvements over the baseline regular decoding.  
While SD-XL-v1.0~\citep{podell2024sdxl} provides marginally better performance, we adopt SD-v1.5 as the default due to its substantially faster image generation time, making it more practical for large-scale or real-time applications.

%% file: tabs/pope.tex
\begin{table*}[t]
    \renewcommand{\arraystretch}{0.93}
    \centering
    \small
    \caption{
        \RebuttalRevision{\textbf{Results on POPE~\citep{li2023evaluating} benchmark}. Higher ($\uparrow$) accuracy, precision, recall, and F1 indicate better performance. The best results are \textbf{bolded}, and the second-best are \underline{underlined}.}
    }
    \label{tab:POPE}
    \setlength{\tabcolsep}{5pt} 
    \resizebox{\textwidth}{!}{
    \begin{tabular}{cclccccccccc}
    \toprule
     & \multirow{2}{*}[-2pt]{\textbf{Setup}} & \multirow{2}{*}[-2pt]{\textbf{Method}} & \multicolumn{3}{c}{\textbf{LLaVA-1.5}} & \multicolumn{3}{c}{\textbf{InstructBLIP}} & \multicolumn{3}{c}{\RebuttalRevision{\textbf{Qwen-VL}}} \\
    \arrayrulecolor{gray} \cmidrule(lr){4-6} \cmidrule(lr){7-9} \cmidrule(lr){10-12}
     &  &  & {Acc.} $\uparrow$ & {Prec.}  $\uparrow$ & {F1} $\uparrow$ & {Acc.} $\uparrow$ & {Prec.} $\uparrow$ & {F1} $\uparrow$ & \RebuttalRevision{{Acc.} $\uparrow$} & \RebuttalRevision{{Prec.} $\uparrow$} & \RebuttalRevision{{F1} $\uparrow$} \\
    \midrule
    \multirow{15}{*}[-5pt]{\rotatebox{90}{\textbf{\normalsize MS-COCO}}} & \multirow{5}{*}{Random} 
    & Regular & 83.13 & 81.94  & 83.44 & 83.07 & 83.02  & 83.08 & 87.43 & 93.56 & 86.48 \\
     &  & VCD  & 87.00 & 86.13  & 87.15 & 86.23 & 88.14  & 85.88 & 88.80 & 93.89 & 88.11 \\
     &  & M3ID  &  87.50 &  87.38  &  87.52 &  86.67 &  88.09  &  86.41 & \underline{89.83} & 95.44 & \underline{89.17}  \\
     &  & RITUAL  &  88.87	&  \underline{89.23} &   88.81 &  88.83 &  90.48  &  88.60 & 89.47 & \underline{96.32} & 88.62 \\
     &  & DeGF  &  \underline{89.23}	& \textbf{90.17}&   \underline{89.11} &  \underline{89.30} &  \underline{90.68}  & \underline{89.10} & 89.73 & 93.19 & \textbf{89.31} \\
     &  & \cc \textbf{Ours} &\cc \textbf{90.33} &\cc 89.05  &\cc \textbf{89.38}  &\cc \textbf{90.12}  &\cc \textbf{91.23}  &\cc \textbf{91.38}  & \cc \textbf{90.11} & \cc \textbf{96.87} & \cc 88.92 \\
     \arrayrulecolor{gray}\cmidrule(lr){2-12}
      &  \multirow{5}{*}{Popular} & Regular & 81.17 &	78.28  &	82.08 & 77.00	& 73.82  &	78.44 & 84.70 & 88.24 & 83.96 \\
     &  & VCD  & 83.10 & 79.96  & 83.94 &  80.07 &  77.67 & 80.89 & 85.13 & 87.27 & 84.69 \\
     &  & M3ID  &  84.30 &  81.58  &  84.95 & 80.97 & 77.93  &  81.85 & 86.27 & \underline{89.19} & \underline{85.73} \\
     &  & RITUAL  & 85.83 &  84.17  &  86.17 &  81.97 &  78.90  &  82.87  & 84.57 & 84.09 & 84.67 \\
     &  &DeGF  &  \underline{86.1}	&  \underline{84.73} &   \underline{86.37} &  \underline{82.50} &  \underline{79.64}  &  \underline{83.31} & \textbf{86.50} & 86.87 & 85.71 \\
     &  & \cc \textbf{Ours} &\cc \textbf{87.67} &\cc \textbf{85.25}  &\cc \textbf{88.43} &\cc \textbf{83.52} &\cc \textbf{80.12} &\cc \textbf{83.69} & \cc \underline{86.34} & \cc \textbf{91.24} & \cc \textbf{86.38} \\
     \arrayrulecolor{gray}\cmidrule(lr){2-12}
      &  \multirow{5}{*}{Adversarial} & Regular & 77.43 & 73.31  & 79.26 & 74.60 & 71.26  & 76.45 & 79.83 & 80.13 & 79.73 \\
     &  & VCD  & 77.17 & 72.18  & 79.47 &  77.20 &  74.29  &  78.49 & 81.33 & 80.60 & 81.55 \\
     &  & M3ID  &  78.23 &  73.51  &  80.22 & 77.47 & 73.68  & 79.14 & 82.03 & 81.47 & 82.19 \\
     &  & RITUAL  &  78.80 &  74.43  &  80.54 &  78.73 &  74.57  &  \textbf{80.39}  & 82.80 & 83.15 & 82.71 \\
     &  & DeGF  &  \underline{79.47}	&  \underline{75.14} &   \underline{81.09} &  \underline{78.8} &  \textbf{78.43}  &  \underline{80.11} & \underline{83.47} & \underline{84.49} & \underline{82.98} \\
     &  & \cc \textbf{Ours} &\cc \textbf{81.67} &\cc \textbf{76.93}   &\cc \textbf{82.22} &\cc \textbf{79.61} &\cc \underline{77.64}  &\cc 79.86 &\cc \textbf{84.36} & \cc \textbf{85.03} & \cc \textbf{83.69} \\
     \arrayrulecolor{gray}\midrule
     \multirow{15}{*}[-5pt]{\rotatebox{90}{\textbf{\normalsize A-OKVQA}}} &  \multirow{5}{*}{Random} & Regular & 81.90 & 76.63  & 83.53 & 80.63 & 76.82  & 81.92 & 86.27 & 90.66 & 85.48 \\
     &  & VCD  & 83.83 & 78.05  & 85.34 & 84.20 &  80.90  & 85.00 & 87.87 & 90.06 & 87.53 \\
     &  & M3ID  &  84.67 &  79.25  &  85.97 &  85.43 & 81.77  &  86.23 & \underline{88.13} & \underline{92.06} & 87.55 \\
     &  & RITUAL  &  85.17 &  79.79  &  86.40 &  87.13 & 83.92 &  87.71 & 87.73 & \textbf{92.49} & 87.01 \\
     &  & DeGF  &  \underline{86.17}	&  \underline{80.84} &   \underline{87.27} &  \underline{87.4} &  \underline{84.67}  &  \underline{88.02} & 87.90 & 89.16 & \underline{87.58} \\
     &  & \cc \textbf{Ours} &\cc \textbf{88.67} &\cc \textbf{83.63}  &\cc \textbf{89.38} &\cc \textbf{88.94} &\cc \textbf{85.32}  &\cc \textbf{89.21} & \cc \textbf{88.33} & \cc 91.46 & \cc \textbf{88.31} \\
     \arrayrulecolor{gray}\cmidrule(lr){2-12}
      &  \multirow{5}{*}{Popular} & Regular & 75.07 &  68.58 & 78.77 & 75.17 & 70.15  & 77.91 & 84.60 & 87.99 & 83.88 \\
     &  & VCD  & 76.63 & 69.59 & 80.19 &  78.63 &  73.53 &  80.72 & 86.23 & 87.30 & 86.03 \\
     &  & M3ID  &  77.80 & 70.98  &  80.91 & 78.80 & 73.38 & 81.00 & \underline{86.50} & 89.59 & 85.95 \\
     &  & RITUAL  &  78.83 &  71.99 &  \underline{81.68} &  {78.73} &  {72.83} &  81.17 & 86.36 & 88.73 & 86.20 \\
     &  & DeGF  &  \underline{79.07}	&  \underline{72.11} &   81.09 &  \underline{80.47} &  \underline{75.61}  &  \underline{82.35} & 86.47 & \underline{90.74} & \underline{86.52} \\
     &  & \cc \textbf{Ours} &\cc \textbf{80.11} &\cc \textbf{72.64} &\cc \textbf{82.36}  &\cc \textbf{80.79} &\cc \textbf{76.29}  &\cc \textbf{83.76} & \cc \textbf{87.71} & \cc \textbf{90.96} & \cc \textbf{87.26} \\
     \arrayrulecolor{gray}\cmidrule(lr){2-12}
      &  \multirow{5}{*}{Adversarial} & Regular & 67.23 & 61.56 & 73.70 & 69.87 & 64.54 & 74.54 & 76.90 & 75.59 & 77.48 \\
     &  & VCD  & 67.40 & 61.39  & 74.21 &  71.00 &  65.41  &  75.45 & 79.13 & 76.04 & 80.30 \\
     &  & M3ID  &  68.60 &  62.22  &  75.11 & 70.10 &  64.28  & 75.16 & 79.50 & 77.54 & 80.21 \\
     &  & RITUAL  &  {68.57} &  62.26  &  {74.99} &  {70.27} & 64.15 & 75.55 & 80.20 & 79.08 & \underline{80.58} \\
     &  & DeGF  &  \underline{70.7}	&  \underline{66.7} &   \underline{76.86} &  \underline{71.87} &  \underline{65.65}  &  \underline{75.96} & \underline{80.75} & \underline{80.37} & 80.46 \\
     &  & \cc \textbf{Ours} &\cc \textbf{71.3} &\cc \textbf{68.1}  &\cc \textbf{78.26} &\cc \textbf{73.44} &\cc \textbf{66.97}   &\cc \textbf{77.16} & \cc \textbf{81.26} & \cc \textbf{80.97} & \cc \textbf{81.04} \\
    \bottomrule
    \end{tabular}
    }
    \vspace{-5pt}
\end{table*}

%% file: sections/5_conclusion.tex
\section{Conclusion}

In this work, we present a multi-granularity generative feedback framework to mitigate hallucination in large vision-language models. Inspired by the hierarchical nature of human visual perception, our method integrates the original image with both coarse-grained global context and fine-grained local visual evidence, enabling a structured multi-resolution reasoning process. By leveraging a generative self-verification loop, our approach allows the model to refine its output based on internal visual imagination and cross-granularity consistency. Experimental results demonstrate that our method significantly improves response faithfulness and visual grounding across multiple benchmarks, particularly in scenarios involving ambiguous or detail-intensive visual inputs. Our framework is model-agnostic, modular, and compatible with existing LVLMs, offering a generalizable pathway toward more reliable multimodal generation. Future work includes exploring adaptive granularity selection via tree-structured exploration and incorporating learned feedback quality estimation to further enhance self-correction capabilities.

\section{Acknowledgments}
This work was supported by the Science and
Technology Innovation 2030-Key Project under Grant 2021ZD0201404.

\clearpage

%% file: sections/6_appendix.tex
\clearpage
\appendix
\renewcommand{\thesection}{\Alph{section}}
\renewcommand\thefigure{\Alph{section}\arabic{figure}} 
\renewcommand\thetable{\Alph{section}\arabic{table}}  
\setcounter{section}{0}
\setcounter{figure}{0} 
\setcounter{table}{0}

\begin{figure}[t]
    \includegraphics[width=\linewidth]{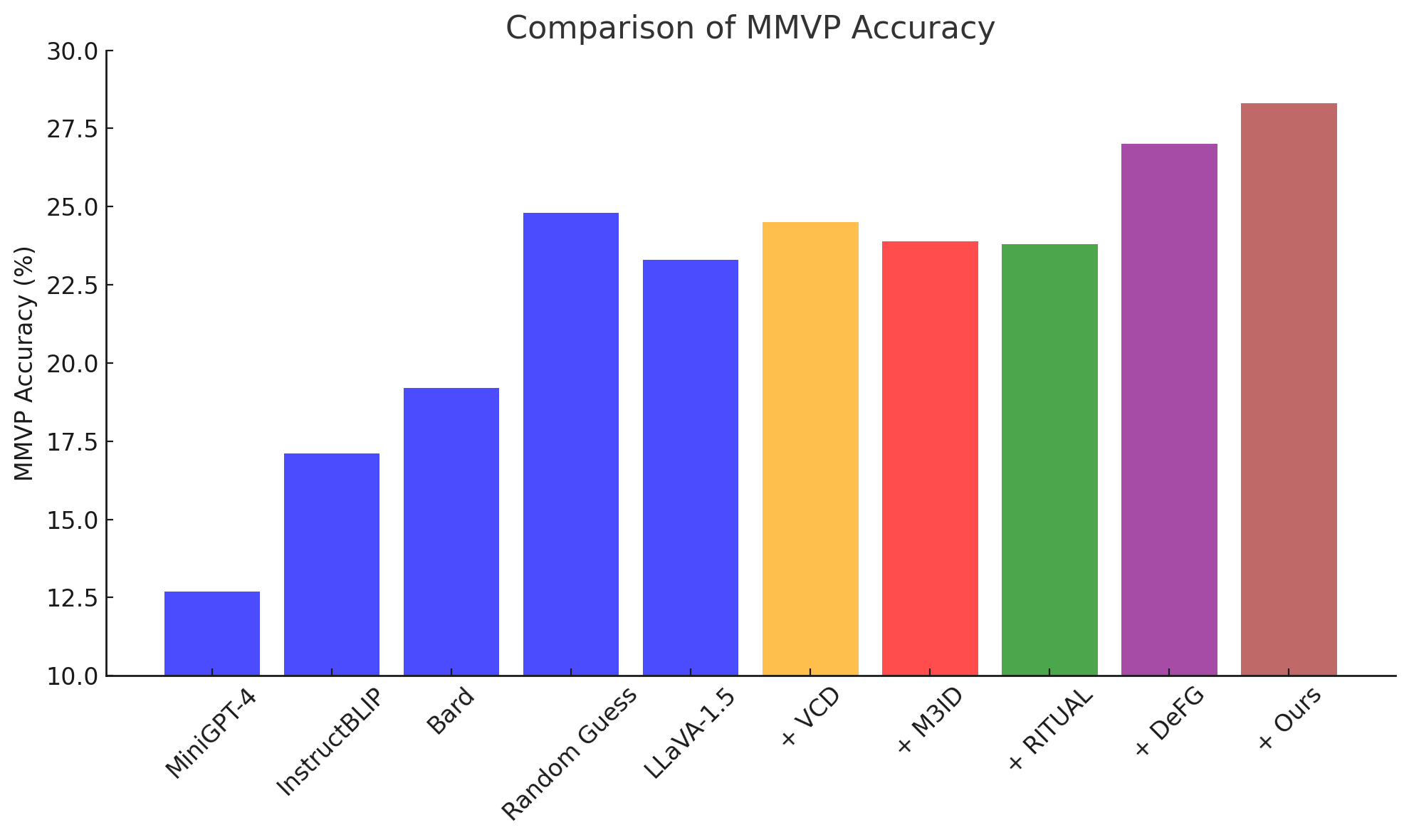}
    \caption{\looseness=-1\textbf{Results on MMVP~\citep{tong2024eyes}}. We apply our approach to LLaVA-1.5~\citep{liu2024improved} and compare its performance against other hallucination mitigation methods.}
    \label{fig:mmvp}
    \end{figure}

\section{More Experimental Results and Analysis}
\label{sec:moreablation}

\begin{table}[t]
    \small
    \begin{center}
        \caption{GPT-4V-aided evaluation on LLaVA-Bench}. Higher accuracy and detailedness ($\uparrow$) indicate better performance. The evaluation is performed on LLaVA-1.5~\citep{liu2024improved}.
        \label{table:gpt4v}
    \resizebox{0.9\linewidth}{!}{
    \begin{tabular}{lcccc}
        \toprule
          \multirow{2}{*}[-0.5ex]{\textbf{Method}}  &  \multicolumn{2}{c}{\textbf{LLaVA-1.5}} & \multicolumn{2}{c}{\textbf{InstructBLIP}} \\
         \cmidrule(lr){2-3}\cmidrule(lr){4-5}
             & Acc. $\uparrow$ & Det. $\uparrow$ & Acc. $\uparrow$ & Det. $\uparrow$ \\
         \midrule
         Regular & 2.88 & 3.29 & 3.42 & 3.96 \\
         DeFG & 4.29 & 4.54  & \cc \textbf{4.38} & 4.79\\
         \cc \textbf{Ours} & \cc \textbf{4.33} & \cc \textbf{4.57}  & 4.33 & \cc \textbf{4.86} \\
         \midrule
          VCD & 3.62 & 3.83  & 3.71 & 4.21\\
          DeFG & 4.04 & \cc \textbf{4.38}  & 4.17 & 4.58\\
          \cc \textbf{Ours} & \cc \textbf{4.12} & 4.31  & \cc \textbf{4.26} & \cc \textbf{4.63} \\
          \midrule
          M3ID & 3.88 & 4.08 & 4.00 & 4.33 \\
          DeFG & 4.04 & \cc \textbf{4.29} & 4.08 & 4.50 \\
          \cc \textbf{Ours} & \cc \textbf{4.13} & 4.18  & \cc \textbf{4.27} & \cc \textbf{4.51} \\
        \bottomrule
    \end{tabular}
    }
    \end{center}
    \end{table}

\noindent\textbf{Results on MMVP}.  
To further evaluate the model's ability in fine-grained visual discrimination, we run experiments on the MMVP benchmark. As illustrated in Figure~\ref{fig:mmvp}, applying our self-refining decoding technique to LLaVA-1.5 elevates performance from 22.67\% to 27.33\%. Compared to other hallucination mitigation baselines, our method yields more substantial improvements, particularly in tasks that require distinguishing visually similar images with subtle differences. These results highlight our approach's ability to improve precision in visual recognition by reducing misinterpretations and hallucinated features, thus delivering more accurate outputs in nuanced visual tasks.

\noindent\textbf{Results on LLaVA-Bench}.  
Figure~\ref{fig:llavabench} showcases a qualitative comparison using LLaVA-Bench, where we examine the responses generated by standard decoding and our CoFi-Dec-enhanced decoding with LLaVA-1.5. Regular decoding often produces vague or hallucinated descriptions, such as references to ``\texttt{the island below the mountain}'' or general observations like ``\texttt{cloudy sky}'' and ``\texttt{cohesive landscape}'', which lack specificity. In contrast, our method produces more grounded and informative outputs, accurately identifying elements such as the volcano, the road, nearby vegetation, and residential areas. GPT-4V-assisted evaluation results, summarized in Table~\ref{table:gpt4v}, further validate these observations, showing that our method outperforms baselines like VCD and M3ID in both response accuracy and descriptive detail.

\begin{table}
  \small
  \begin{center}
  \caption{\textbf{Efficiency comparison}. For each method, we present the average inference latency per instance and peak GPU memory. Experiments are conducted on a single RTX A6000 Ada GPU.}
  \label{tab:efficiency}
  \vspace{-5pt}
  \resizebox{\linewidth}{!}{
  \begin{tabular}{lcccc}
  \toprule
  Method  & Avg. Latency $\downarrow$ & GPU Memory $\downarrow$ & CHAIR$_S$ $\downarrow$  \\ 
  \midrule
  Regular  &  3.44 s {\tiny ($\times$1.00)} &  15778 MB  {\tiny ($\times$1.00)} & 55.0\\
  VCD &  6.91 s {\tiny ($\times$2.01)} &  16634 MB  {\tiny ($\times$1.05)}  & 54.4 \\
  OPERA & 24.70 s  {\tiny ($\times$7.18)} &  22706 MB  {\tiny ($\times$1.44)}& 52.6 \\
  Woodpecker & 10.68 s {\tiny ($\times$3.10)} & 22199 MB  {\tiny ($\times$1.41)} & 57.6 \\
  HALC & 22.61 s {\tiny ($\times$6.51)} &  23084 MB {\tiny ($\times$1.46)}& 51.0 \\
  DeFG &  13.89 s {\tiny ($\times$4.04)}  & 19119 MB  {\tiny ($\times$1.21)}& 48.8\\
  \rowcolor{gray!20} \textbf{Ours} &  20.83 s {\tiny ($\times$6.05)}  & 18597 MB  {\tiny ($\times$1.28)}& 46.7\\
  \bottomrule
  \end{tabular}
  }
  \end{center}
  \end{table}

  \input{tabs/ablation}
  \noindent\textbf{Efficiency Comparison.}
  In Table~\ref{tab:efficiency}, we present a comparative analysis of the computational efficiency of our method versus other baseline approaches on the CHAIR benchmark, utilizing the LLaVA-1.5 model with a maximum sequence length of 128 tokens. Our method requires two forward passes and integrates a text-to-image generation step to suppress hallucinations, which leads to a latency increase of approximately 4.04$\times$ and a GPU memory overhead of 1.21$\times$ relative to standard decoding.  
  The full inference process in our framework consists of three sequential phases: (1) initial response generation, (2) visual feedback generation via a diffusion model, and (3) refinement of the original response. On average, these stages take 3.4s, 3.8s, and 6.6s per sample, respectively. Although our method is less efficient than simpler techniques like standard or contrastive decoding, it is notably more efficient than computationally intensive approaches such as OPERA and HALC.  
  Importantly, our framework consistently achieves the lowest hallucination rates across all evaluated methods.

\begin{table*}[t]
    \renewcommand{\arraystretch}{0.93}
    \centering
    \small
    \caption{
        \RebuttalRevision{\textbf{Another results on POPE~\citep{li2023evaluating} benchmark}. Higher ($\uparrow$) accuracy, precision, recall, and F1 indicate better performance. The best results are \textbf{bolded}, and the second-best are \underline{underlined}.}
    }
    \label{tab:POPE-another}
    \setlength{\tabcolsep}{5pt} 
    \resizebox{\textwidth}{!}{
    \begin{tabular}{cclccccccccc}
    \toprule
     & \multirow{2}{*}[-2pt]{\textbf{Setup}} & \multirow{2}{*}[-2pt]{\textbf{Method}} & \multicolumn{3}{c}{\textbf{LLaVA-1.5}} & \multicolumn{3}{c}{\textbf{InstructBLIP}} & \multicolumn{3}{c}{\RebuttalRevision{\textbf{Qwen-VL}}} \\
    \arrayrulecolor{gray} \cmidrule(lr){4-6} \cmidrule(lr){7-9} \cmidrule(lr){10-12}
     &  &  & {Acc.} $\uparrow$ & {Prec.}  $\uparrow$ & {F1} $\uparrow$ & {Acc.} $\uparrow$ & {Prec.} $\uparrow$ & {F1} $\uparrow$ & \RebuttalRevision{{Acc.} $\uparrow$} & \RebuttalRevision{{Prec.} $\uparrow$} & \RebuttalRevision{{F1} $\uparrow$} \\
    \midrule
     \multirow{15}{*}[-5pt]{\rotatebox{90}{\textbf{\normalsize GQA}}} &  \multirow{5}{*}{Random} & Regular & 82.23 & 76.32  & 84.03 & 79.67 & 76.05  & 80.99 & 84.90 & 89.51 & 83.96 \\
     &  & VCD  & 83.23 & 76.73  & 85.05 &  82.83 &  80.16  &  83.56 & 85.21 & 92.05 & 84.21 \\
     &  & M3ID  &  84.20 &  78.00  &  85.77 & 83.07 & 80.06 & 83.87 & 85.69 & 93.11 & 84.67 \\
     &  & RITUAL  &  86.10 &  {80.30} &  {87.31} &  {84.87} &  {82.52}  &  \underline{85.39} & \underline{86.1} & {93.78} & {84.81} \\
     &  & DeGF  &  \underline{87.09}	&  \underline{80.46} &   \underline{87.96} &  \underline{85.40} &  \underline{85.64}  &  85.12 & 85.95 & \underline{94.22} & \underline{85.08} \\
     &  & \cc \textbf{Ours} &\cc \textbf{89.03} &\cc \textbf{81.1}   &\cc \textbf{89.06} &\cc \textbf{86.78} &\cc \textbf{87.06}  &\cc \textbf{86.39} & \cc \textbf{87.14} & \cc \textbf{94.65} & \cc \textbf{86.32} \\
     \arrayrulecolor{gray}\cmidrule(lr){2-12}
      &  \multirow{5}{*}{Popular} & Regular & 73.47 & 66.83  & 77.84 & 73.33 & 68.72  & 76.26 & 81.33 & 83.38 & 80.74 \\
     &  & VCD  & 72.37 & 65.27 & 77.58 & \underline{76.13} & 71.10  & \textbf{78.68} & 81.97 & 82.82 & 81.73 \\
     &  & M3ID  &  73.87 &  66.70  &  78.49 &  75.17 &  69.94  &  78.04 & 82.13 & 84.58 & 81.48 \\
     &  & RITUAL  &  74.80 &  67.50  &  79.15 &  {74.50} &  {69.17}  &  {77.61} & 81.13 & 85.48 & 81.03 \\
     &  & DeGF  &  \underline{75.12}	&  \underline{71.56} &   \underline{80.98} &  75.34 &  \underline{71.89}  &  77.96 & \underline{82.10} & \underline{86.39} & \underline{81.85} \\
     &  & \cc \textbf{Ours} &\cc \textbf{78.56} &\cc \textbf{73.28}   &\cc \textbf{83.34}  &\cc \textbf{76.18} &\cc \textbf{73.17} &\cc \underline{78.65} & \cc \textbf{83.54} & \cc \textbf{87.24} & \cc \textbf{83.09} \\
     \arrayrulecolor{gray}\cmidrule(lr){2-12}
      &  \multirow{5}{*}{Adversarial} & Regular & 68.60 & 62.43  & 74.84 &  68.60 &  63.94  &  73.10 & 79.03 & 80.43 & 78.54 \\
     &  & VCD  & 68.83 & 62.26  & 75.43 & 71.00 & 65.75  & 75.14 & 80.87 & 81.07 & 80.80 \\
     &  & M3ID  &  68.67 &  62.16  &  75.28 & 71.17 & 65.79 & \underline{75.36} & 81.03 & 82.93 & \underline{80.94} \\
     &  & RITUAL  &  {68.23} &  {61.75} &  {75.10} &  {70.17} &  {64.76}  &  {74.78} & 81.07 & 83.29 & 80.41 \\
     &  & DeGF  &  \underline{74.07}	&  \underline{67.42} &   \underline{78.22} &  \underline{72.45} &  \underline{68.52}  &  {75.32} & \underline{81.13} & \underline{84.18} & 80.57 \\
     &  & \cc \textbf{Ours} &\cc \textbf{75.21} &\cc \textbf{68.34}  &\cc \textbf{79.76} &\cc \textbf{73.25} &\cc \textbf{69.68}  &\cc \textbf{76.87} & \cc \textbf{81.69} & \cc \textbf{84.96} & \cc \  \textbf{81.67}\\
    \bottomrule
    \end{tabular}
    }
    \vspace{-5pt}
\end{table*}



    \begin{figure*}[t]
        \centering
        \includegraphics[width=0.9\linewidth]{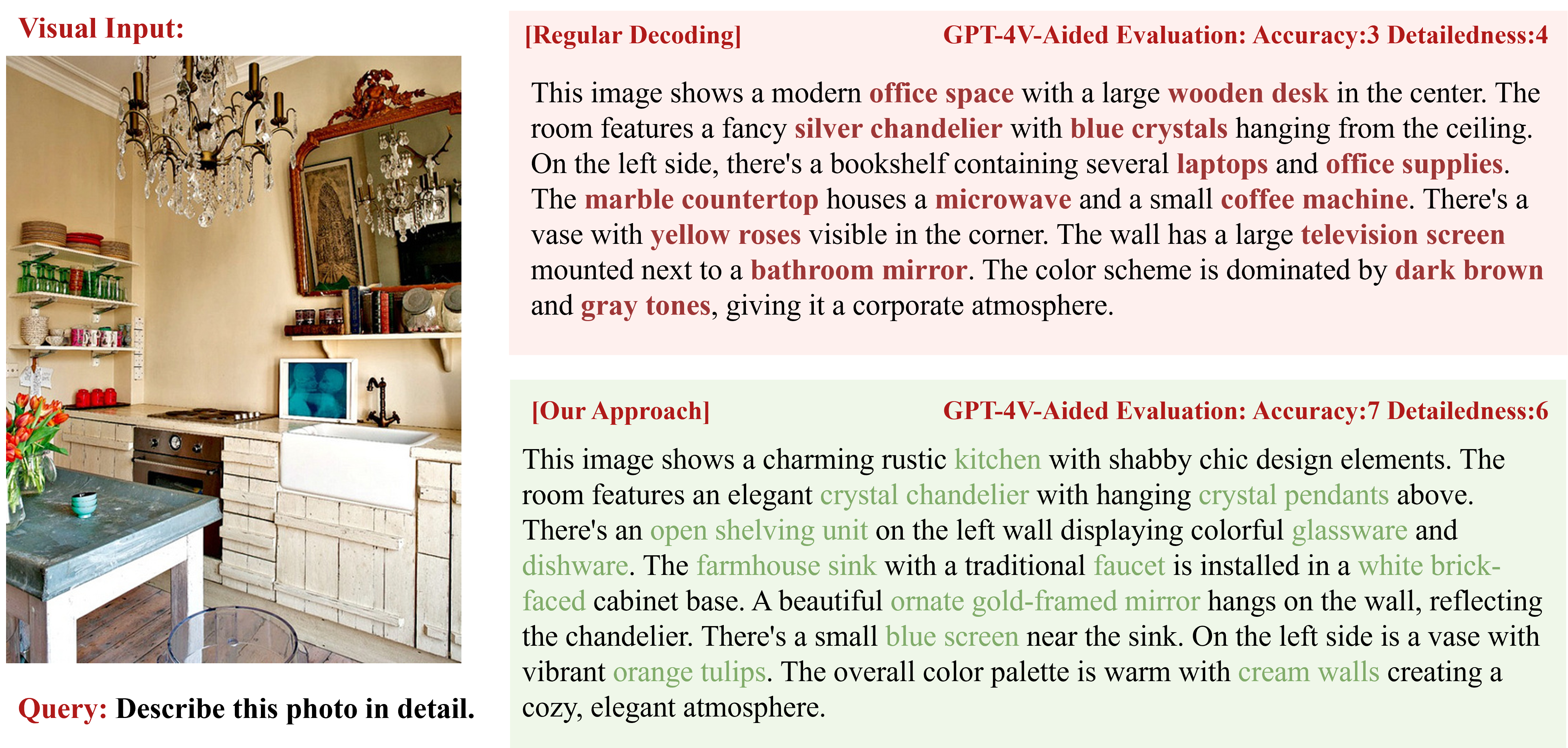}
        \caption{Case study on the LLaVA-Bench benchmark. We compare the responses generated by regular decoding and our method using LLaVA-1.5. GPT-4V-aided evaluation results are also provided alongside the responses. Hallucinated and accurate content is highlighted in \textcolor{darkred}{red} and \textcolor{darkgreen}{green}.} 
        \label{fig:case1}
        \end{figure*}

\begin{figure*}[t]
    \centering
    \includegraphics[width=0.9\linewidth]{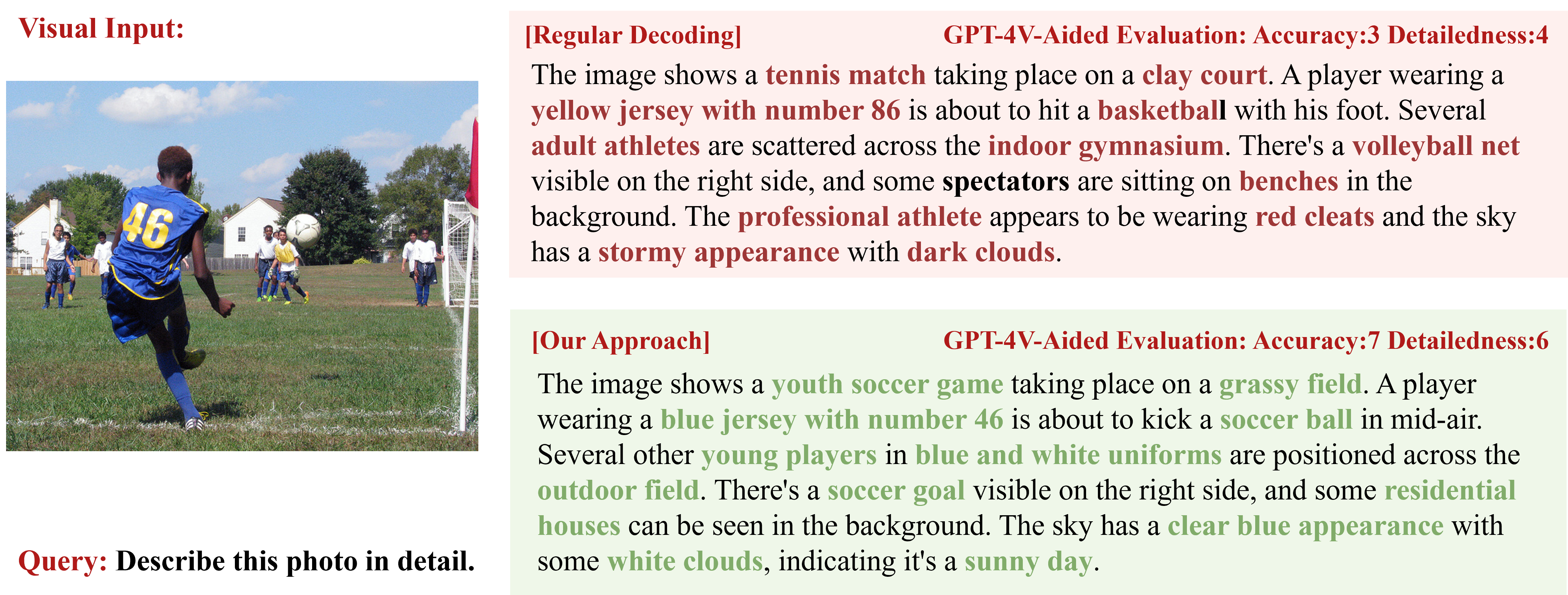}
    \caption{Case study on the LLaVA-Bench benchmark.} 
    \label{fig:case3}
    \end{figure*}

    \begin{figure*}[t]
        \centering
        \includegraphics[width=0.9\linewidth]{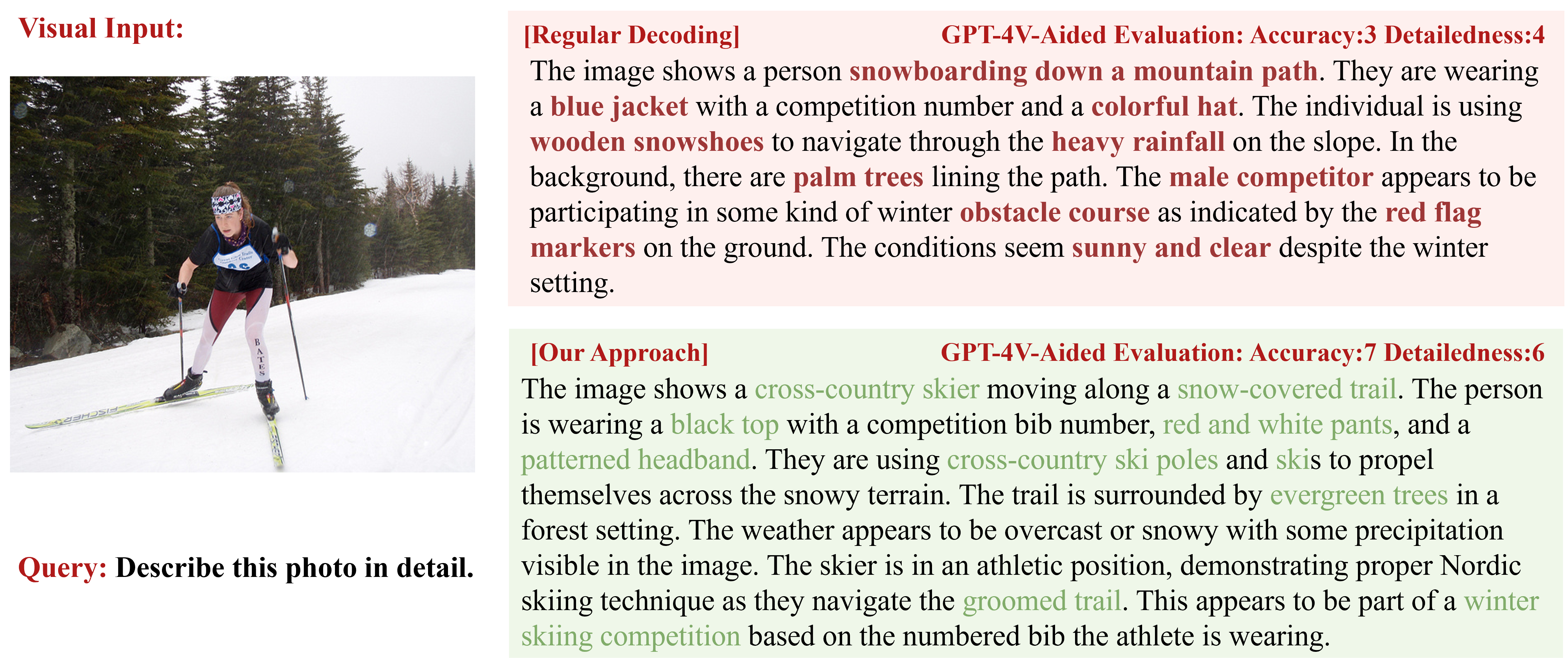}
        \caption{Case study on the LLaVA-Bench benchmark.} 
        \label{fig:case4}
        \end{figure*}

\begin{figure*}[t]
    \centering
    \includegraphics[width=0.9\linewidth]{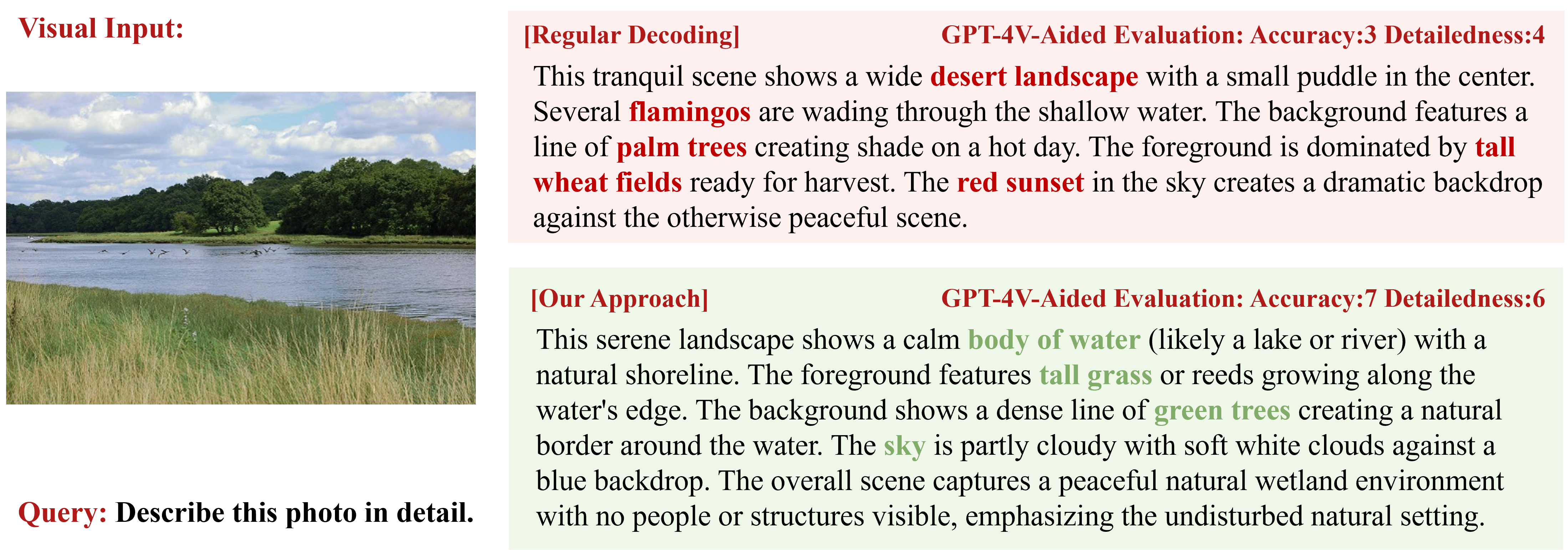}
    \caption{Case study on the LLaVA-Bench benchmark.} 
    \label{fig:case5}
    \end{figure*}

\begin{figure*}[t]
    \centering
    \includegraphics[width=0.9\linewidth]{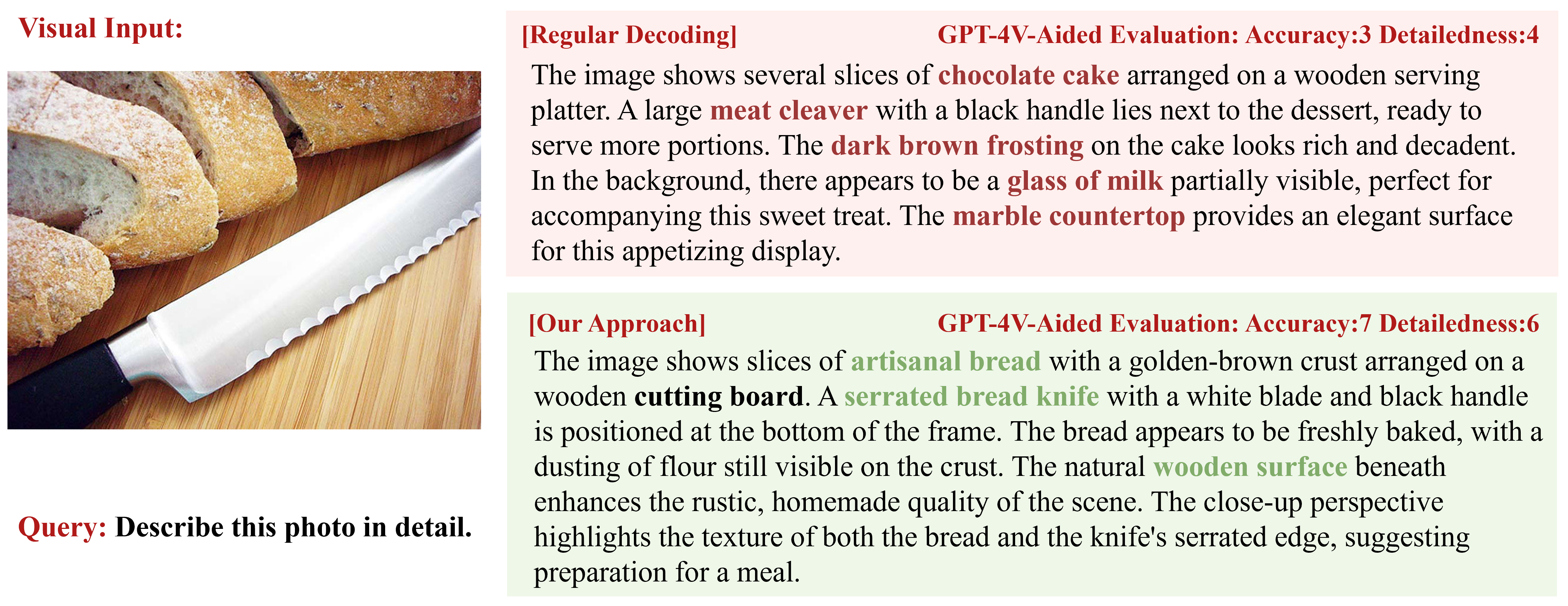}
    \caption{Case study on the LLaVA-Bench benchmark.} 
    \label{fig:case6}
    \end{figure*}

\begin{figure*}[t]
    \centering
    \includegraphics[width=0.9\linewidth]{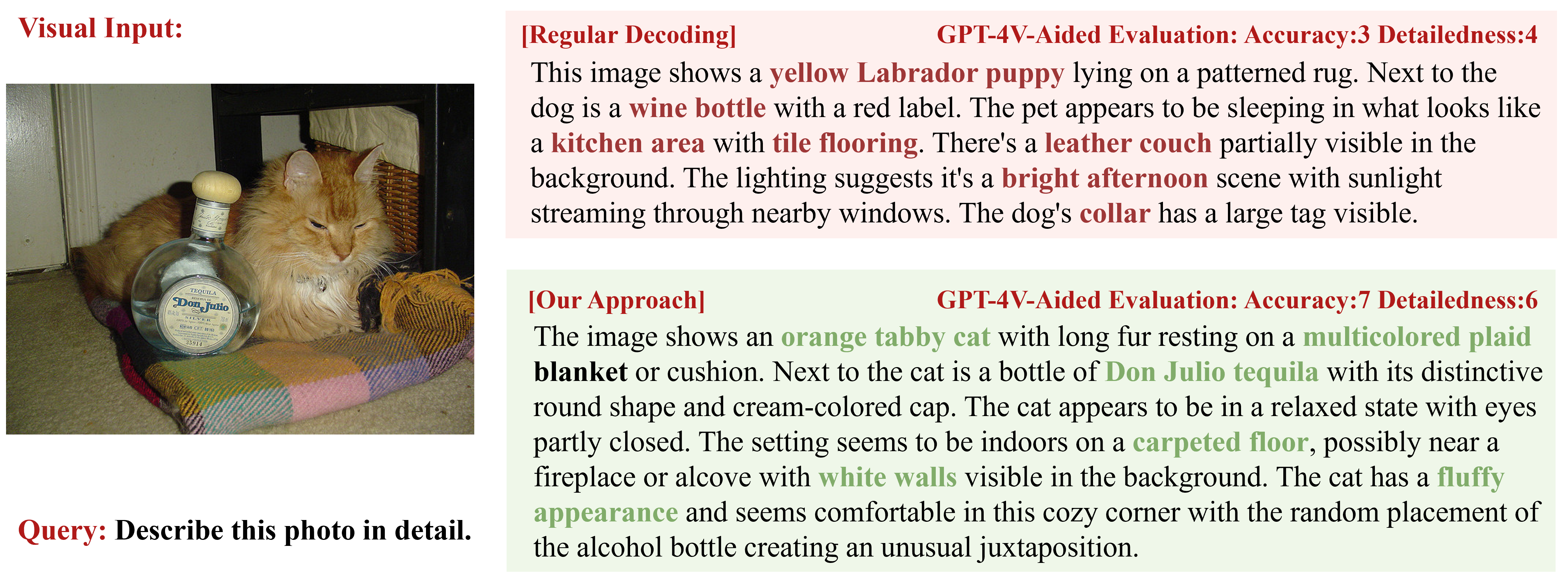}
    \caption{Case study on the LLaVA-Bench benchmark.} 
    \label{fig:case7}
    \end{figure*}

    \begin{figure*}[t]
        \centering
        \includegraphics[width=0.9\linewidth]{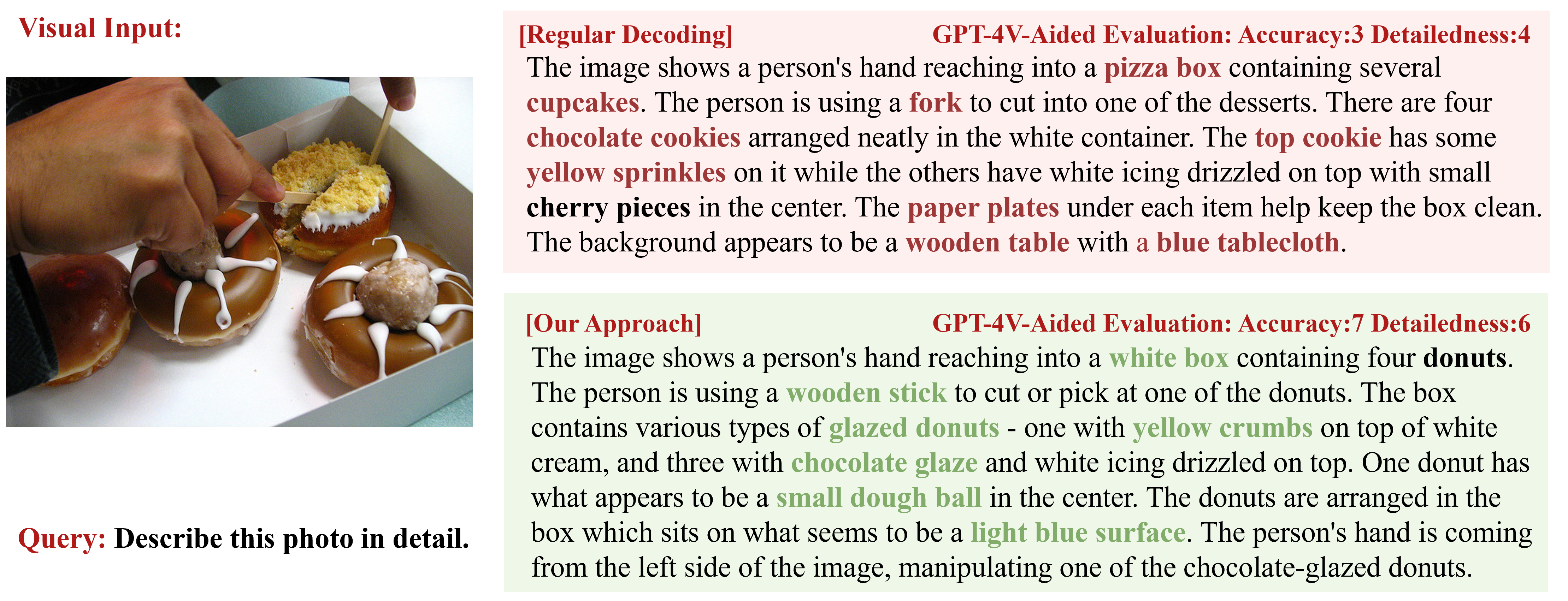}
        \caption{Case study on the LLaVA-Bench benchmark.} 
        \label{fig:case9}
        \end{figure*}

        \begin{figure*}[t]
            \centering
            \includegraphics[width=0.9\linewidth]{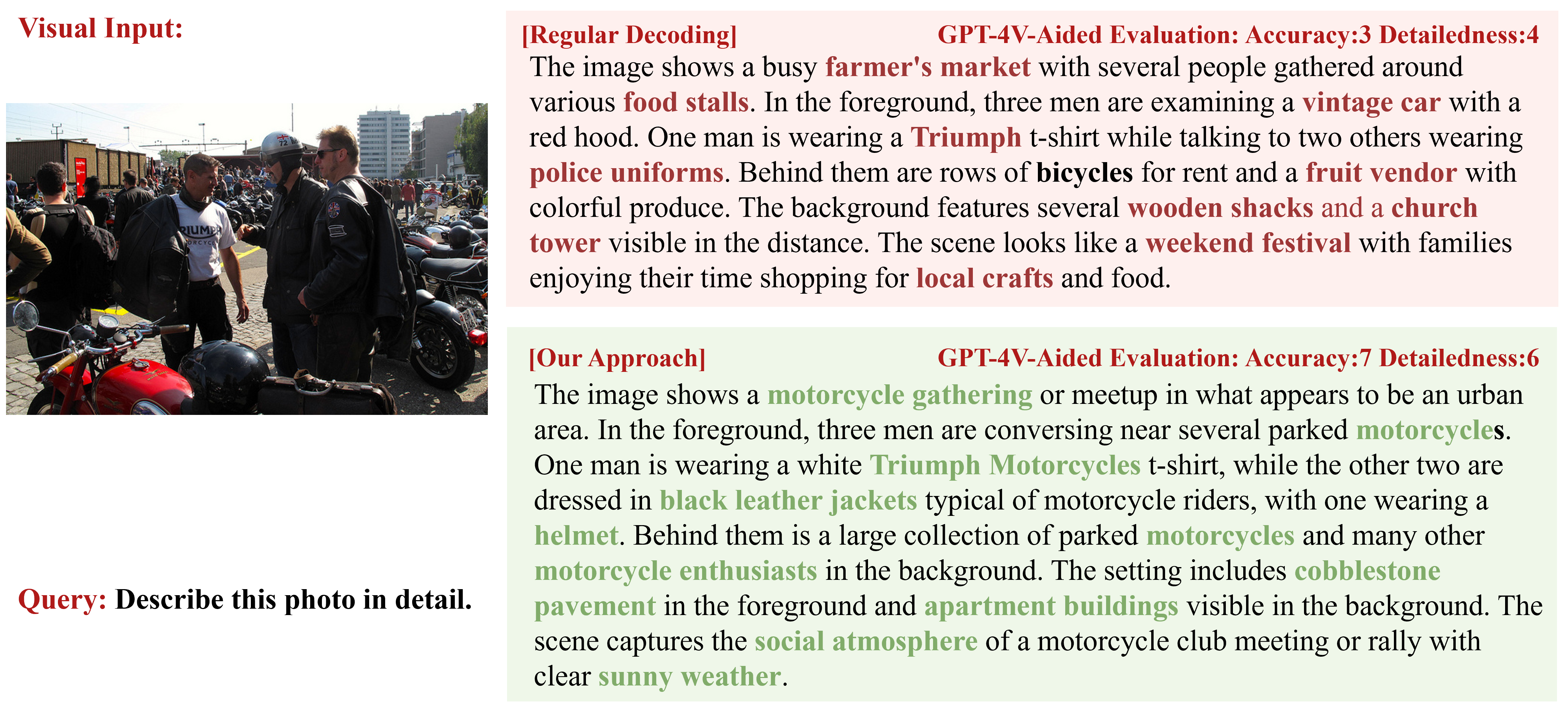}
            \caption{Case study on the LLaVA-Bench benchmark.} 
            \label{fig:case10}
            \end{figure*}

%% file: tabs/ablation.tex
\begin{figure}[t]
\begin{minipage}[t]{0.95\linewidth}
\makeatletter\def\@captype{table}
\setlength\tabcolsep{2.5pt}
\caption{\textbf{Effects of different generative models.} We report the performance of different variants of our method, utilizing various stable diffusion models, on the LLaVA-1.5 backbone.}
    \vspace{1pt}
    \label{tab:generative}
    \resizebox{\textwidth}{!}{
    \begin{tabular}{lcccc}
        \toprule
        Models & POPE Acc. & CHAIR$_S$ & CHAIR$_I$ & MME Score \\
        \midrule
        Regular & 83.13 & 26.2 & 9.4 & 562.50  \\
        SD-v1.1 & 88.37 & 19.3 & 6.5 & 638.33 \\
        \cc SD-v1.5 & \cc\textbf{89.03} & \cc18.4 & \cc6.1 & \cc644.44 \\
        SD-v2.1 & 88.70 & 18.8 & 6.7 & 632.22 \\
        SD-XL-v0.9 & 88.87 & 18.6 & 6.1 & 642.50 \\
        SD-XL-v1.0 & 88.60 & \textbf{17.9} & \textbf{5.8} & \textbf{648.33} \\
        \bottomrule
    \end{tabular}
}
\end{minipage}
\end{figure}